\documentclass[11pt,reqno]{amsart}
\usepackage[margin=1in]{geometry}
\usepackage[foot]{amsaddr}
\usepackage{longtable}
\usepackage{algorithm}
\usepackage{algorithmic}
\usepackage{graphicx}
\usepackage{comment}
\usepackage{multirow}
\usepackage{pdfpages}

\usepackage{amsmath, amsthm, amssymb, url, color, hyperref, bbm}
\hypersetup{
    colorlinks,
    citecolor=blue,
    filecolor=black,
    linkcolor=black,
    urlcolor=black
}

\newcommand{\R}{\mathbb{R}}
\newcommand{\C}{\mathbb{C}}
\newcommand{\Z}{\mathbb{Z}}
\renewcommand{\P}{\mathbb{P}}
\newcommand{\E}{\mathbb{E}}
\newcommand{\x}{\mathbf{x}}
\newcommand{\y}{\mathbf{y}}
\newcommand{\z}{\mathbf{z}}
\renewcommand{\a}{\mathbf{a}}
\newcommand{\e}{\mathbf{e}}
\renewcommand{\u}{\mathbf{u}}
\renewcommand{\v}{\mathbf{v}}

\renewcommand{\r}{\mathbf{r}}
\newcommand{\A}{\mathbf{A}}
\newcommand{\U}{\mathbf{U}}
\newcommand{\M}{\mathbf{M}}
\newcommand{\bP}{\mathbf{P}}
\newcommand{\Q}{\mathbf{Q}}
\newcommand{\T}{\mathsf{T}}
\newcommand{\N}{\mathcal{N}}
\newcommand{\F}{\mathcal{F}}
\newcommand{\1}{\mathbf{1}}
\newcommand{\I}{\mathbf{I}}
\newcommand{\0}{\mathbf{0}}
\renewcommand{\S}{\mathcal{S}}

\newcommand{\op}{\operatorname{op}}
\newcommand{\TV}{\mathrm{TV}}
\newcommand{\ITALE}{\mathrm{ITALE}}

\newcommand{\sign}{\operatorname{sign}}
\DeclareMathOperator*{\argmin}{arg\,min}

\newcommand{\bSigma}{\mathbf{\Sigma}}
\renewcommand{\i}{\mathbf{i}}

\newtheorem{theorem}{Theorem}[section]

\newtheorem{lemma}[theorem]{Lemma}
\newtheorem{proposition}[theorem]{Proposition}
\theoremstyle{definition}
\newtheorem{definition}[theorem]{Definition}

\begin{document}

\title[ITALE for estimating gradient-sparse signals]{Iterative Alpha Expansion for estimating gradient-sparse signals from linear measurements}

\author{Sheng Xu}
\email{sheng.xu@yale.edu}

\author{Zhou Fan}
\email{zhou.fan@yale.edu}
\address{Department of Statistics and Data Science \\ Yale University \\
New Haven, CT 06511}

\maketitle

\begin{abstract}
We consider estimating a piecewise-constant image, or a gradient-sparse signal
    on a general graph, from noisy linear measurements. We propose and study an
    iterative algorithm to minimize a penalized least-squares objective, with a
    penalty given by the ``$\ell_0$-norm'' of the signal's discrete graph
    gradient. The method proceeds by approximate proximal descent, applying the
    alpha-expansion procedure to minimize a proximal gradient in each iteration,
    and using a geometric decay of the penalty parameter across iterations.
    Under a cut-restricted isometry property for the measurement design, we
    prove global recovery guarantees for the estimated signal. For standard
    Gaussian designs, the required number of measurements is independent of the
    graph structure, and improves upon worst-case guarantees for
    total-variation (TV) compressed sensing on the 1-D and 2-D lattice graphs by
    polynomial and logarithmic factors, respectively. The method empirically
    yields lower mean-squared recovery error compared with TV regularization in
    regimes of moderate undersampling and moderate to high signal-to-noise,
    for several examples of changepoint signals and gradient-sparse phantom images.
\end{abstract}

\section{Introduction}

Consider an unknown signal $\x_* \in \R^p$ observed via
$n$ noisy linear measurements
\[\y=\A\x_*+\e \in \R^n.\]
We study the problem of estimating $\x_*$, under the assumption that
its coordinates correspond to the $p$ vertices of a given
graph $G=(V,E)$, and $\x_*$ is gradient-sparse. By this, we mean that
\begin{equation}\label{eq:l0norm}
\|\nabla \x_*\|_0 \equiv \sum_{(i,j) \in E} \1\{x_{*,i} \neq x_{*,j}\}
\end{equation}
is much smaller than the total number of edges $|E|$. Special cases of interest
include the 1-D line graph, where variables have a sequential order and $\x_*$
has a changepoint structure, and the 2-D lattice graph, where coordinates of
$\x_*$ represent pixels of a piecewise-constant image.

This problem has been studied since early pioneering works in compressed
sensing \cite{candesrobust,candesstable,donoho}. 
Among widely-used approaches for estimating $\x_*$ are
those based on constraining or penalizing the total-variation (TV)
semi-norm \cite{rudinosherfatemi}, which may be defined (anisotropically) for a general graph as
\[\|\nabla \x\|_1 \equiv \sum_{(i,j) \in E} |x_i-x_j|.\]
These are examples of $\ell_1$-analysis methods
\cite{eladetal,candescoherent,nametal}, which regularize the $\ell_1$-norm of a
general linear transform of $\x$ rather than of its coefficients in an
orthonormal basis.
Related fused-lasso methods have been studied for different applications of
regression and prediction in \cite{tibshiranifused,rinaldo,rjtibshirani},
other graph-based regularization methods for linear regression in
\cite{lietal,kimgao}, and trend-filtering methods 
regularizing higher-order discrete derivatives of $\x$ in
\cite{kimetal,wangetal}.

Theoretical recovery guarantees for TV-regularization depend on the structure
of the graph \cite{needellward,needellward3d,caixu}, and more generally on
sparse conditioning properties of the pseudo-inverse $\nabla^\dagger$ for
$\ell_1$-analysis methods with sparsifying transform $\nabla$. For direct
measurements $\A=\I$, these and related issues were studied in
\cite{hutterrigollet,dalalyanetal,fanguan}, which showed in particular
that TV-regularization may not achieve the same worst-case
recovery guarantees as analogous $\ell_0$-regularization methods on
certain graphs including the 1-D line. In this setting of
$\A=\I$, different computational approaches exist which may be used for
approximately minimizing an $\ell_0$-regularized objective on general graphs
\cite{boykovetal,kleinbergtardos,xuetal}.

Motivated by this line of work, we propose and study
an alternative to TV-regularization for the problem with
indirect linear measurements $\A \neq \I$. Our procedure is based similarly on
the idea of minimizing a possibly non-convex objective
\begin{equation}\label{eq:generalobjective}
F(\x)=\frac{1}{2}\|\y-\A\x\|_2^2+\lambda \sum_{(i,j) \in E} c(x_i,x_j)
\end{equation}
for an edge-associated cost function $c$. We will focus attention in this work
on the specific choice of an $\ell_0$-regularizer
\begin{equation}\label{eq:l0cost}
c(x_i,x_j)=\1\{x_i \neq x_j\},
\end{equation}
which matches (\ref{eq:l0norm}), although the algorithm may be
applied with more general choices of metric edge cost. For the above $\ell_0$ edge
cost, the resulting objective takes the form
\[F(\x)=\frac{1}{2}\|\y-\A\x\|_2^2+\lambda \|\nabla \x\|_0.\]

We propose to minimize $F(\x)$ using an iterative algorithm akin to proximal
gradient descent: For parameters $\gamma \in (0,1)$ and $\eta>0$,
the iterate $\x_{k+1}$ is computed from $\x_k$ via
\begin{align*}
\a_{k+1} &\leftarrow \x_k-\eta \A^\T(\A\x_k-\y)\\
\x_{k+1} ``&\leftarrow" \argmin_{\x}
\frac{1}{2}\|\x-\a_{k+1}\|_2^2+\lambda_k \sum_{(i,j) \in E} c(x_i,x_j)\\
\lambda_{k+1} &\leftarrow \lambda_k \cdot \gamma
\end{align*}
For general graphs, the second update step for $\x_{k+1}$ is only approximately
computable in polynomial time. We apply the alpha-expansion procedure of Boykov, Veksler, and Zabih
\cite{boykovetal} for this task, first discretizing the continuous signal
domain, as analyzed statistically in \cite{fanguan}. In contrast to analogous
proximal methods in convex settings \cite{beckteboulle,parikhboyd}, where
typically $\lambda_k \equiv \lambda\eta$ is fixed across iterations, we decay $\lambda_k$ geometrically from a large initial value to ensure algorithm convergence. We call the resulting algorithm ITerative ALpha Expansion, or ITALE.

Despite $F(\x)$ being non-convex and non-smooth, we provide global recovery
guarantees for a suitably chosen ITALE iterate $\x_k$. For example,
under exact gradient-sparsity $\|\nabla \x_*\|_0=s_*$, if $\A$ consists of
\begin{equation}\label{eq:ndependence}
n \gtrsim s_*\log (1+|E|/s_*)
\end{equation}
linear measurements with i.i.d.\ $\N(0,1/n)$ entries, then the ITALE iterate $\x_k$ for the $\ell_0$-regularizer (\ref{eq:l0cost}) and a penalty value $\lambda_k \asymp \|\e\|_2^2/s_*$ satisfies with high probability
\begin{equation}\label{eq:heuristicguarantee}
\|\x_k-\x_*\|_2 \lesssim \|\e\|_2.
\end{equation}
More generally, we provide recovery guarantees when $\A$ satisfies a certain
cut-restricted isometry property, described in Definition \ref{def:RIP} below.
(In accordance with
the compressed sensing literature, we state all theoretical guarantees for
deterministic and possibly adversarial measurement error $\e$.)

Even for i.i.d.\ Gaussian design,
we are not aware of previous polynomial-time algorithms which provably achieve
this guarantees for either the 1-D line or
the 2-D lattice. In particular, connecting with the above discussion, similar
existing results for TV-regularization in noisy or noiseless settings require $n
\gtrsim s_*(\log |E|)^3$ Gaussian measurements for the 2-D lattice and $n
\gtrsim \sqrt{|E|s_*}\log |E|$ measurements for the 1-D line
\cite{needellward,caixu}. In contrast, for lattice graphs of dimensions 3 and higher where
the Laplacian $\mathbf{L}=\nabla^\T\nabla$ is well-conditioned, as
well as for more general $\ell_1$-analysis methods where $\nabla^\dagger$ is a
tight frame, optimal recovery guarantees for TV/$\ell_1$-regularization hold
with $n \gtrsim s_* \log |E|$ or $n \gtrsim s_* \log (|E|/s_*)$ measurements as
expected \cite{candescoherent,needellward3d,caixu}. ITALE provides this
guarantee up to a constant factor, irrespective of the graph structure.

\begin{figure}[t]
\begin{minipage}{0.33\textwidth}
\includegraphics[width=0.9\textwidth]{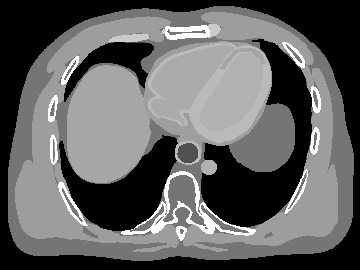}
\end{minipage}%
\begin{minipage}{0.65\textwidth}
\includegraphics[width=0.45\textwidth]{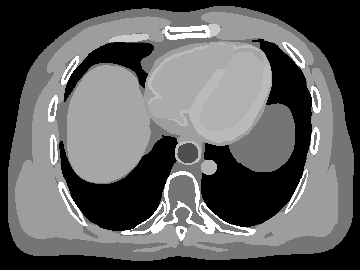}%
\includegraphics[width=0.45\textwidth]{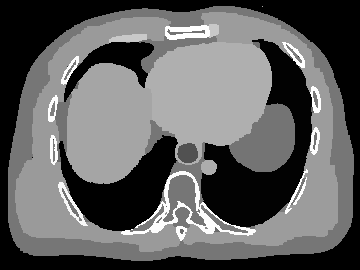}\\
\includegraphics[width=0.45\textwidth]{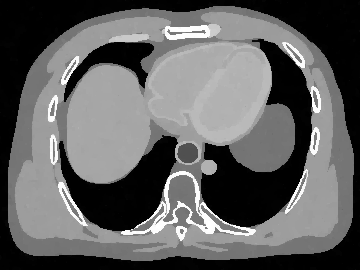}%
\includegraphics[width=0.45\textwidth]{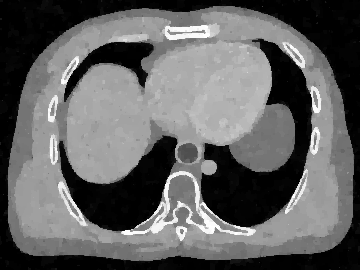}%
\end{minipage}
\caption{Left: Original image slice from the XCAT digital phantom. Top row:
$\hat{\x}^\ITALE$ from 20\% undersampled and reweighted Fourier measurements, in low noise ($\sigma=4$, left) and medium noise ($\sigma=16$, right) settings. Bottom row: $\hat{\x}^\TV$ for the same measurements.}\label{fig:XCAT}
\end{figure}

In practice, for $\gamma$ sufficiently close to 1, we directly interpret the
sequence of ITALE iterates $\x_k$ as approximate
minimizers of the objective function (\ref{eq:generalobjective}) for penalty
parameters $\lambda=\lambda_k/\eta$
along a regularization path. We select the iterate $k$ using cross-validation on the prediction error for $\y$, and we use the final estimate $\hat{\x}^\ITALE=\x_k$.
Figure \ref{fig:XCAT} compares in simulation $\hat{\x}^\ITALE$ using the
$\ell_0$-regularizer (\ref{eq:l0cost}) with $\hat{\x}^\TV$ (globally) minimizing
the TV-regularized objective
\begin{equation}\label{eq:TVobjective}
F^\TV(\x)=\frac{1}{2}\|\y-\A\x\|_2^2+\lambda\|\nabla \x\|_1,
\end{equation}
with $\lambda$ selected also using cross-validation.
The example depicts a synthetic image of a human chest slice,
previously generated by \cite{gongetal} using the XCAT digital phantom
\cite{segarsetal}. The design $\A$ is an undersampled and reweighted Fourier
matrix, using a sampling scheme described in Section \ref{sec:theory} and
similar to that proposed in \cite{krahmerward} for TV-regularized compressed
sensing. In a low-noise setting, a detailed comparison of the recovered images
reveals that $\hat{\x}^\ITALE$ provides a sharper
reconstruction than $\hat{\x}^\TV$. As noise increases, $\hat{\x}^\TV$
becomes blotchy, while $\hat{\x}^\ITALE$ begins to lose finer image
details. Quantitative comparisons of recovery error are provided in
Section \ref{sec:2D} and are favorable towards ITALE in lower noise regimes.

ITALE is similar to some methods oriented towards $\ell_0$-regularized
sparse regression and signal recovery \cite{troppgilbert,zhang,bertsimasetal},
including notably the Iterative Hard Thresholding (IHT)
\cite{blumensathdavies} and CoSaMP \cite{needelltropp} methods in compressed
sensing. We highlight here several differences:
\begin{itemize}
\item For sparsity in an orthonormal basis, forward stepwise selection and orthogonal matching pursuit provide greedy ``$\ell_0$'' approaches to variable selection, also with provable guarantees \cite{troppgilbert,zhang,elenbergetal}.
However, such methods do not have direct analogues for
gradient-sparsity in graphs, as one cannot select a single edge
difference $x_i-x_j$ to be nonzero without changing other edge differences.
\item IHT and CoSaMP enforce sparsity of $\x_{k+1}$ in each iteration by
projecting to the $s$ largest coordinates of $\a_{k+1}$, for user-specified $s$.
In contrast, ITALE uses a Lagrangian form that penalizes (rather than
constrains) $\|\nabla \x_{k+1}\|_0$. This is partly for computational reasons, as we are not aware of fast algorithms
that can directly perform such a projection step onto the (non-convex) set
$\{\x:\|\nabla \x\|_0 \leq s\}$ for general graphs. Our theoretical convergence analysis handles this Lagrangian form.
\item In contrast to more general-purpose mixed-integer optimization procedures
in \cite{bertsimasetal}, each iterate of ITALE (and hence also the full algorithm, for a polynomial number of iterations) is provably polynomial-time
in the input size $(n,p,|E|)$ \cite{fanguan}. On our personal computer, for the
$p=360 \times 270=97200$ image of Figure \ref{fig:XCAT}, computing the 60
iterates constituting a full ITALE solution path required about 20 minutes, using the optimized alpha-expansion code of \cite{boykovkolmogorov}.
\end{itemize}

While our theoretical focus is on $\ell_0$-regularization, we expect that
for certain regimes of undersampling and signal-to-noise,
improved empirical recovery may be possible with edge costs $c(x_i,x_j)$
interpolating between the $\ell_0$ and $\ell_1$ penalties. These are applicable
in the ITALE algorithm and would be interesting to investigate in future work.

\section{Model and algorithm}\label{sec:model}
Let $G=(V,E)$ be a given connected graph on the vertices
$V=\{1,\ldots,p\}$, with undirected edge set $E$. We assume throughout that $p \geq 3$. For a signal vector $\x_* \in \R^p$, measurement matrix $\A \in \R^{n \times p}$, and measurement errors $\e \in \R^n$, we observe
\begin{equation}\label{eq:model}
    \y=\A\x_*+\e \in \R^n.
\end{equation}
Denote by $\nabla \in \{-1,0,1\}^{|E| \times p}$ the discrete gradient matrix
on the graph $G$, defined\footnote{Here, we may fix an arbitrary ordering of the vertex pair $(i,j)$ for each edge.} by
\[\nabla \x=\big(x_i-x_j:\;(i,j) \in E\big) \in \R^{|E|}.\]
We study estimation of $\x_*$, assuming that $\x_*$ has (or is
well-approximated by a signal having) small exact gradient sparsity $\|\nabla \x_*\|_0$.

Our proposed algorithm is an iterative approach called ITALE, presented as Algorithm \ref{alg:ITALE}. It is based around the idea of minimizing the objective (\ref{eq:generalobjective}). In this objective, the cost function $c:\R^2 \to \R$ must satisfy the metric properties
\begin{equation}\label{eq:cost}
c(x,y)=c(y,x) \geq 0, \qquad c(x,x)=0 \Leftrightarrow x=0, \qquad
c(x,z) \leq c(x,y)+c(y,z),
\end{equation}
but is otherwise general. Importantly, $c$ may be non-smooth and non-convex. 
The algorithm applies proximal descent, alternating between constructing a
surrogate signal $\a_{k+1}$ in line 3 and denoising this surrogate signal in
line 4, discussed in more detail below.

Some intuition for $\a_{k+1}$ is provided by considering the setting $\e \approx \0$ and $\eta\A^\T\A \approx \I$, in which case
\begin{align*}
\a_{k+1}&=\x_k-\eta\A^\T(\A\x_k-\y)\\
&=\x_*+(\I-\eta\A^\T\A)(\x_k-\x_*)+\eta\A^\T\e
\approx \x_*.
\end{align*}
There are two sources of noise $(\I-\eta\A^\T\A)(\x_k-\x_*)$ and $\eta\A^\T\e$ in $\a_{k+1}$, the former expected to decrease across iterations as the reconstruction error $\|\x_k-\x_*\|$ decreases. A tuning parameter $\lambda_k$ is applied to denoise $\a_{k+1}$ in each iteration, where $\lambda_k$ also decreases across iterations to match the noise level. Our theoretical analysis indicates to use a geometric rate of decay $\lambda_{k+1}=\lambda_k \cdot \gamma$, starting from a large initial value $\lambda_{\max}$.

\begin{algorithm}[t]
    \caption{Iterative Alpha Expansion}\label{alg:ITALE}
    \begin{algorithmic}[1]
        \REQUIRE{$\y \in \R^n$, $\A \in \R^{n \times p}$, and parameters
        $\gamma \in (0,1)$, $\lambda_{\max}>\lambda_{\min}>0$, and
        $\eta,\delta>0$.}
    \STATE{Initialize $\x_0 \gets \0$, $\lambda_0 \gets \lambda_{\max}$}
    \FOR{$k=0,1,2,\ldots,K$ until $\lambda_K<\lambda_{\min}$}
        \STATE{$\a_{k+1} \gets \x_k-\eta\A^\T(\A\x_k-\y)$}
        \STATE{$\x_{k+1} \gets
        \text{AlphaExpansion}(\a_{k+1},\lambda_k,\delta)$}
        \STATE{$\lambda_{k+1} \gets \lambda_k \cdot \gamma$}
    \ENDFOR
        \ENSURE{$\x_1,\ldots,\x_K$}
\end{algorithmic}
\end{algorithm}

ITALE yields iterates $\x_1,\x_2,\ldots,\x_K$, which we directly interpret as
recovered signals along a regularization path for different choices of $\lambda
\equiv \lambda_k/\eta$ in the objective (\ref{eq:generalobjective}). We choose
$\lambda_{\max}$ such that the initial iterates oversmooth $\x_*$, and
$\lambda_{\min}$ such that the final iterates undersmooth $\x_*$. We remark that
an alternative approach would be to iterate lines 3 and 4 in Algorithm \ref{alg:ITALE} until convergence for each $\lambda_k$, before updating $\lambda_k$ to the next value $\lambda_{k+1}$. However, we find that this is not necessary in practice if $\gamma$ is chosen close enough to 1, and our stated algorithm achieves a computational speed-up compared to this approach.

To perform the denoising in line 4, ITALE applies the alpha-expansion graph cut procedure from \cite{boykovetal} to approximately solve the minimization problem
\[\min_{\x \in \R^p} \frac{1}{2}\|\x-\a_{k+1}\|_2^2+\lambda_k\sum_{(i,j) \in E} c(x_i,x_j).\]
This sub-routine is denoted as $\text{AlphaExpansion}(\a_{k+1},\lambda_k,\delta)$, and is described in Algorithm \ref{alg:AlphaExpansion} for completeness. 
At a high level, the alpha-expansion method encodes the above objective function in the
structure of an edge-weighted augmented graph, and iterates over global moves
that swap the signal value on a subset of vertices for a given new value by
finding a minimum graph cut. The original alpha-expansion algorithm of
\cite{boykovetal} computes an approximate maximum-a-posteriori estimate in a
discrete Potts model with a metric edge-cost satisfying (\ref{eq:cost}).
To apply this to a continuous signal domain, we restrict coordinate values of $\x$ to a discrete grid
\[\delta\Z=\{k\delta:k \in \Z\}.\]
Here, $\delta$ is a small user-specified discretization parameter.
As shown in \cite[Lemma S2.1]{fanguan} (see also \cite[Theorem
6.1]{boykovetal}), the output $\x_{k+1}=\text{AlphaExpansion}(\a_{k+1},\lambda_k,\delta)$ has the deterministic guarantee
\begin{equation}\label{eq:AEguarantee}
    \frac{1}{2}\|\x_{k+1}-\a_{k+1}\|_2^2+\lambda_k\sum_{(i,j) \in E} c(x_i,x_j)
    \leq \min_{\x \in (\delta\Z)^p} \left(\frac{1}{2}\|\x-\a_{k+1}\|_2^2
    +2\lambda_k\sum_{(i,j) \in E} c(x_i,x_j)\right)
\end{equation}
with the additional factor of 2 applying to the penalty on the right side. This guarantee is important for the theoretical recovery properties that we will establish in Section \ref{sec:theory}.

\begin{algorithm}[t]
\caption{AlphaExpansion$(\a,\lambda,\delta)$ subroutine}\label{alg:AlphaExpansion}
\begin{algorithmic}[1]
\REQUIRE{$\a \in \R^p$, cost function $c:\R^2 \to \R$, parameters $\lambda,\delta>0$.}
\STATE{Let $a_{\min},a_{\max}$ be the minimum
and maximum values of $\a$. Initialize $\x \in \R^p$ arbitrarily.}
\LOOP
  \FOR{each $z \in \delta\Z \cap [a_{\min},a_{\max}]$}
  \STATE{Construct the following edge-weighted augmentation
$G_{z,\x}$ of the graph $G$:}
\begin{ALC@g}
\STATE{Introduce a source vertex $s$ and a sink vertex $t$,
connect $s$ to each $i \in \{1,\ldots,p\}$ with weight
$\frac{1}{2}(a_i-z)^2$, and connect $t$ to each $i \in \{1,\ldots,p\}$ with weight
$\frac{1}{2}(a_i-x_i)^2$ if
$x_i \neq z$, or weight $\infty$ if $x_i=z$.}
\FOR{each edge $\{i,j\} \in E$}
\IF{$x_i=x_j$}
\STATE{Assign weight $\lambda c(x_i,z)$ to $\{i,j\}$.}
\ELSE
\STATE{Introduce a new vertex $v_{i,j}$, and replace edge $\{i,j\}$ by the three edges $\{i,v_{i,j}\}$,
$\{j,v_{i,j}\}$, and $\{t,v_{i,j}\}$, with weights
$\lambda c(x_i,z)$, $\lambda c(x_j,z)$, and $\lambda c(x_i,x_j)$ respectively.}
\ENDIF
\ENDFOR
\end{ALC@g}
     \STATE{Find the minimum s-t cut $(S,T)$ of $G_{z,\x}$ such
that $s \in S$ and $t \in T$.}
\STATE{For each $i \in \{1,\ldots,p\}$, update $x_i \leftarrow z$ if $i \in T$, and keep $x_i$ unchanged if $i \in S$.}
  \ENDFOR
  \STATE{If $\x$ was unchanged for each $z$ above, then return $\x$.}
\ENDLOOP
\ENSURE{$\x$}
\end{algorithmic}
\end{algorithm}

We make a few remarks regarding parameter tuning in practice:
\begin{itemize}
\item Using conservative choices for $\lambda_{\max}$ (large), $\gamma$ (close to 1), and $\delta$ (small) increases the total runtime of the procedure, but does not degrade the quality of recovery. In our experiments, we fix $\gamma=0.9$ and set $\delta$ in each iteration to yield 300 grid values for $\delta \Z \cap [a_{\min},a_{\max}]$ in Algorithm \ref{alg:AlphaExpansion}.
\item We monitor the gradient sparsity $\|\nabla \x_k\|_0$ across iterations, and terminate the algorithm when $\|\nabla \x_K\|_0$ exceeds a certain fraction (e.g.\ 50\%) of the total number of edges $|E|$, rather than fixing $\lambda_{\min}$.
\item The parameter $\eta$ should be matched to the scaling and restricted isometry properties of the design matrix $\A$. For sub-Gaussian and Fourier designs scaled by $1/\sqrt{n}$ as in Propositions \ref{prop:subgaussian} and \ref{prop:2DFourier} below, we set $\eta=1$.
\item The most important tuning parameter is the iterate $k$ for which we take
the final estimate $\hat{\x}^\ITALE=\x_k$. In practice, we apply
cross-validation on the mean-squared prediction error for $\y$ to select $k$.
Note that $\eta$ should be rescaled by the number of training samples in each fold, i.e.\ for 5-fold cross-validation with training sample size $0.8n$, we set $\eta=1/0.8$ instead of $\eta=1$ in the cross-validation runs.
\end{itemize}

\section{Recovery guarantees}\label{sec:theory}

We provide in this section theoretical guarantees on the recovery error
$\|\hat{\x}^\ITALE-\x_*\|_2$, where $\hat{\x}^\ITALE \equiv \x_k$ for a
deterministic (non-adaptive) choice of iterate $k$. Throughout this section, ITALE is assumed to be applied with the $\ell_0$ edge cost $c(x_i,x_j)=\1\{x_i \neq x_j\}$.

\subsection{cRIP condition}\label{sec:RIP}
Our primary assumption on the measurement design $\A$ will be the following version of a restricted isometry property.

\begin{definition}\label{def:RIP}
    Let $\kappa>0$, and let $\rho:[0,\infty) \to [0,\infty)$ be any function satisfying $\rho'(s) \geq 0$ and $\rho''(s) \leq 0$ for all $s>0$. A matrix $\A \in \R^{n \times p}$ satisfies the $(\kappa,\rho)$-{\bf cut-restricted isometry property} (cRIP)
    if, for every $\x \in \R^p$ with $\|\nabla \x\|_0 \geq 1$, we have
    \[\left(1-\kappa-\sqrt{\rho(\|\nabla\x\|_0)}\right)
    \|\x\|_2 \leq \|\A\x\|_2 \leq
    \left(1+\kappa+\sqrt{\rho(\|\nabla\x\|_0)}\right)\|\x\|_2.\]
\end{definition}

This definition depends implicitly on the structure of the underlying graph $G$, via its discrete gradient matrix $\nabla$. Examples of the function $\rho$ are given in the two propositions below.

This condition is stronger than the usual RIP condition in compressed sensing \cite{candesrobust,candesstable} in two ways: First, Definition \ref{def:RIP} requires quantitative control of $\|\A\x\|_2$ for \emph{all} vectors $\x \in \R^p$, rather than only those with sparsity $\|\nabla \x\|_0 \leq s$ for some specified $s$. We use this in our analysis to handle regularization of $\|\nabla \x\|_0$ in Lagrangian (rather than constrained) form. Second, approximate isometry is required for signals with small gradient-sparsity $\|\nabla \x\|_0$, rather than small sparsity $\|\x\|_0$. For graphs with bounded maximum degree, all sparse signals are also gradient-sparse, so this is indeed stronger up to a relabeling of constants. This requirement is similar to the D-RIP condition of \cite{candescoherent} for general sparse analysis models, and is also related to the condition of \cite{needellward} that $\A\mathcal{H}^{-1}$ satisfies the usual RIP condition, where $\mathcal{H}^{-1}$ is the inverse Haar-wavelet transform on the 2-D lattice.

Despite this strengthening of the required RIP condition, Definition \ref{def:RIP} still holds for sub-Gaussian designs $\A$, where $\kappa$ depends on the condition number of the design covariance. We defer the proof of the following result to Appendix \ref{appendix:RIP}. For a random vector $\a$, we denote its sub-Gaussian norm as
\[\|\a\|_{\psi_2}=\sup_{\u:\|\u\|_2=1} \sup_{k \geq 1}\; k^{-1/2}\E\Big[|\u^\T\a|^k\Big]^{1/k},\]
and say that $\a$ is sub-Gaussian if $\|\a\|_{\psi_2} \leq K$ for a constant $K>0$.

\begin{proposition}\label{prop:subgaussian}
Let $\A \in \R^{n\times p}$ have i.i.d.\ rows $\a_i/\sqrt{n}$, where $\operatorname{Cov}[\a_i]=\bSigma$ and $\|\a_i\|_{\psi_2} \leq K$. Suppose that the largest and smallest eigenvalues of $\bSigma$ satisfy $\sigma_{\max}(\bSigma) \leq (1+\kappa)^2$ and $\sigma_{\min}(\bSigma)\geq (1-\kappa)^2$ for a constant $\kappa \in (0,1)$. Then for any $k>0$ and some constant $C>0$ depending only on $K,\kappa,k$, with probability at least $1-|E|^{-k}$, the matrix $\A$ satisfies $(\kappa,\rho)$-cRIP for the function
\[\rho(s)=\frac{Cs\log(1+|E|/s)}{n}.\]
\end{proposition}

For large 2-D images, using Fourier measurements with matrix multiplication
implemented by an FFT can significantly reduce the runtime of Algorithm \ref{alg:ITALE}. As previously discussed in \cite{lustigetal,needellward,krahmerward}, uniform random sampling of Fourier coefficients may not be appropriate for reconstructing piecewise-constant images, as these typically have larger coefficients in the lower Fourier frequencies. We instead study a non-uniform sampling and reweighting scheme similar to that proposed in \cite{krahmerward} for total-variation compressed sensing, and show that Definition \ref{def:RIP} also holds for this reweighted Fourier matrix.

For $p=N_1N_2$ and $N_1,N_2$ both powers of 2, let $\F \in \C^{p \times p}$ be the 2-D discrete Fourier matrix on the lattice graph $G$ of size $N_1 \times N_2$, normalized such that $\F\F^*=\I$. We define this as the Kronecker product $\F=\F^1 \otimes \F^2$, where $\F^1 \in \C^{N_1 \times N_1}$ is the 1-D discrete Fourier matrix with entries
\[\F^1_{jk}=\frac{1}{\sqrt{N_1}}
\cdot e^{2\pi \i \cdot \frac{(j-1)(k-1)}{N_1}},\]
and $\F^2 \in \C^{N_2 \times N_2}$ is defined analogously.
(Thus rows closer to $N_1/2+1$ in $\F^1$ correspond to higher frequency components.) Let $\F_{(i,j)}^*$ denote row $(i,j)$ of $\F$, where we index by pairs
$(i,j) \in \{1,\ldots,N_1\} \times \{1,\ldots,N_2\}$ corresponding to the Kronecker structure. We define a sampled Fourier matrix as follows: Let $\nu_1$ be the probability mass function
on $\{1,\ldots,N_1\}$ given by
\begin{equation}\label{eq:nu}
\nu_1(i) \propto \frac{1}{C_0+\min(i-1,N_1-i+1)}, \qquad C_0 \geq 1.
\end{equation}
Define similarly $\nu_2$ on $\{1,\ldots,N_2\}$, and let $\nu=\nu_1 \times \nu_2$.
For a given number of measurements $n$, draw $(i_1,j_1),\ldots,(i_n,j_n) \overset{iid}{\sim} \nu$, and set
\begin{equation}\label{eq:weightedfourier}
    \tilde{\A}=\frac{1}{\sqrt{n}}\begin{pmatrix} \F_{(i_1,j_1)}^*/\sqrt{\nu(i_1,j_1)}
   \\ \vdots \\ \F_{(i_n,j_n)}^*/\sqrt{\nu(i_n,j_n)}
\end{pmatrix} \in \C^{n \times p}.
\end{equation}

\begin{proposition}\label{prop:2DFourier}
    Let $G$ be the 2-D lattice graph of size $N_1 \times N_2$, where $N_1,N_2$ are powers of 2 and $1/K<N_1/N_2<K$ for a constant $K>0$. Set $p=N_1N_2$ and let $\tilde{\A}$ be the matrix defined in (\ref{eq:weightedfourier}). Then for some constants
    $C,t_0>0$ depending only on $K$, and for any $t>t_0$, with probability at least $1-e^{-(\log n) (\log p)^3}-p^{-t}$, $\tilde{\A}$
    satisfies the $(\kappa,\rho)$-cRIP with $\kappa=0$ and
    \[\rho(s)=Cts\frac{(\log p)^8\log n}{n}.\]
\end{proposition}

We defer the proof also to Appendix \ref{appendix:RIP}. This proposition pertains to the complex analogue of Definition \ref{def:RIP}, where $\tilde{\A},\x$ are allowed to be complex-valued, and $\|\cdot\|_2$ denotes the complex $\ell_2$-norm. For a real-valued signal $\x_* \in \R^p$, Algorithm \ref{alg:ITALE} may be applied to $\tilde{\y}=\tilde{\A}\x_*+\e \in \C^n$ by separating real and imaginary parts of $\tilde{\y}$ into a real vector $\y \in \R^{2n}$. The corresponding $\A \in \R^{2n \times p}$ satisfies $\|\A\x\|^2=\|\tilde{\A}\x\|^2$, so the same cRIP condition holds (in the real sense) for $\A$.

\subsection{Recovery error bounds}

To illustrate the idea of analysis, we first establish a result showing that ITALE can yield exact recovery in a setting of no measurement noise. We require $\x_*$ to be gradient-sparse with coordinates belonging exactly to $\delta\Z$, as the ITALE output has this latter property. Discretization error will be addressed in our subsequent result.

\begin{theorem}\label{thm:noiseless}
Suppose $\e=\0$ and $\x_* \in (\delta\Z)^p$, and denote $s_*=\max(\|\nabla \x_*\|_0,1)$.
Suppose $\sqrt{\eta} \cdot \A$ satisfies $(\kappa,\rho)$-cRIP, where $\kappa \in [0,\sqrt{3/2}-1)$. Set $t(\kappa)=1-4\kappa-2\kappa^2 \in (0,1]$, and choose tuning parameters
\[\left(1-\frac{t(\kappa)}{4}\right)^2<\gamma<1,
\qquad \lambda_{\max}>\|\x_*\|_2^2.\]
For some constants $C,c>0$ depending only on $\kappa$, if
\[\rho(s_*) \leq c,\]
then each iterate $\x_k$ of Algorithm \ref{alg:ITALE} satisfies
    \begin{equation}
        \|\x_k-\x_*\|_2 \leq C\sqrt{\lambda_{\max}s_*} \cdot \gamma^{k/2}.
        \label{eq:recoverynonoise}
    \end{equation}
In particular, $\x_k=\x_*$ for all sufficiently large $k$.
\end{theorem}

Thus, in this noiseless setting, the iterates exhibit linear convergence to the true signal $\x_*$. The required condition $\rho(s_*) \leq c$ translates into a requirement of
\[n \gtrsim s_*\log(1+|E|/s_*)\]
measurements for $\A$ having i.i.d.\ $\N(0,1/n)$ entries, by Proposition \ref{prop:subgaussian}, or
\[n \gtrsim s_*(\log p)^8\log\log p\]
weighted Fourier measurements for the 2-D lattice graph, as defined in Proposition \ref{prop:2DFourier}. For these designs, $(\kappa,\rho)$-cRIP holds for $\sqrt{\eta} \cdot \A$ where $\kappa=0$ and $\eta=1$.

\begin{proof}[Proof of Theorem \ref{thm:noiseless}]
    Denote
    \[s_k=\|\nabla \x_k\|_0, \qquad \r_k=\x_k-\x_*.\]
    Applying the optimality condition (\ref{eq:AEguarantee}) to compare
    $\x_{k+1}$ with $\x_*=\x_k-\r_k$, we obtain
    \begin{equation}\label{eq:noiselessoptimality}
        \|\x_{k+1}-\a_{k+1}\|_2^2+2\lambda_ks_{k+1}
        \leq \|\x_k-\r_k-\a_{k+1}\|_2^2+4\lambda_ks_*.
    \end{equation}

    Let $\S_k$ be the partition of $\{1,\ldots,p\}$ induced by the
    piecewise-constant structure of $\x_k$: Each element of $\S_k$ corresponds
    to a connected subgraph of $G$ on which $\x_k$ takes a constant value.
    Let $\S_{k+1},\S_*$ similarly be the partitions induced by
    $\x_{k+1},\x_*$, and denote by $\S$ the common refinement of
    $\S_k,\S_{k+1},\S_*$. Defining the boundary
    \[\partial \S=\{(i,j) \in E:\;i,j \text{ belong to different elements of }
    \S\},\]
    observe that each edge $(i,j) \in \partial S$ must be such that at least
    one of $\x_k$, $\x_{k+1}$, or $\x_*$ takes different values at its two
    endpoints. Then
    \begin{equation}\label{eq:boundarysize}
        |\partial \S| \leq s_*+s_k+s_{k+1}.
    \end{equation}
    
    Let $\bP:\R^p \to \R^p$ be the orthogonal projection
    onto the subspace of signals taking a constant value over each
    element of $\S$, and let $\bP^\perp=\I-\bP$. Then $\x_{k+1},\x_k,\r_k$ all
    belong to the range of $\bP$, so an orthogonal decomposition yields
    \begin{align*}
        \|\x_{k+1}-\a_{k+1}\|_2^2&=
        \|\x_{k+1}-\bP\a_{k+1}\|_2^2+\|\bP^\perp\a_{k+1}\|_2^2,\\
        \|\x_k-\r_k-\a_{k+1}\|_2^2
        &=\|\x_k-\r_k-\bP\a_{k+1}\|_2^2
        +\|\bP^\perp \a_{k+1}\|_2^2.
    \end{align*}
    Applying this, the definition (in the noiseless setting $\e=\0$)
    \[\a_{k+1}=\x_k-\eta \A^\T(\A\x_k-\y)=\x_k-\eta\A^\T\A\r_k,\]
    and the condition $\bP\x_k=\x_k$ to (\ref{eq:noiselessoptimality}),
    we obtain
        \[\|\x_{k+1}-\x_k+\eta\bP\A^\T\A\r_k\|_2^2
        \leq \|\eta\bP\A^\T\A\r_k-\r_k\|_2^2
        +\lambda_k(4s_*-2s_{k+1}).\]
    Applying the triangle inequality and $\x_{k+1}-\x_k=\r_{k+1}-\r_k$,
    \begin{equation}\label{eq:tmpbound}
        \left(\|\r_{k+1}\|_2-\|\r_k-\eta\bP\A^\T\A\r_k\|_2\right)_+^2
    \leq \|\r_k-\eta\bP\A^\T\A\r_k\|_2^2+\lambda_k(4s_*-2s_{k+1}).
    \end{equation}

    We derive from this two consequences: First, lower-bounding the left side
    by 0 and rearranging,
\begin{equation}\label{eq:tmpbound2}
       \lambda_k s_{k+1} \leq \frac{1}{2}\|\r_k-\eta\bP\A^\T\A\r_k\|_2^2+ 2\lambda_k s_*
       \leq \|\r_k\|_2^2+\|\sqrt{\eta}\A\bP\|_{\op}^2  \cdot\|\sqrt{\eta}\A\r_k\|_2^2+ 2\lambda_k s_*.
\end{equation}         
    The condition (\ref{eq:boundarysize}) and definition of $\bP$ imply, for any $\u \in
    \R^p$, that $\|\nabla(\bP\u)\|_0 \leq s_*+s_k+s_{k+1}$. The definition of $\r_k$ implies $\|\nabla \r_k\|_0\leq s_*+s_k$.
    Setting
    \[\tau_k=\kappa+\sqrt{\rho(s_*+s_k+s_{k+1})},\qquad \zeta_k=\kappa+\sqrt{\rho(s_*+s_k)}\]
    we deduce from the $(\kappa,\rho)$-cRIP condition for $\sqrt{\eta} \cdot \A$
    that
    \begin{equation}\label{eq:tmpbound3}
    \|\sqrt{\eta}\A\bP\|_{\op}^2=\sup_{\u\in\R^p:\|\u\|_2=1}\|\sqrt{\eta}\A\bP\u\|_2^2\leq (1+\tau_k)^2, \qquad
    \|\sqrt{\eta}\A\r_k\|_2^2\leq (1+\zeta_k)^2\|\r_k\|_2^2.
    \end{equation}
    Note that since $\rho(s)$ and $\sqrt{\rho(s)}$ are both nonnegative and
concave by Definition \ref{def:RIP}, we have
    \[\rho'(s) \leq (\rho(s)-\rho(0))/s \leq \rho(s)/s,
    \qquad \frac{d}{ds}[\sqrt{\rho(s)}] \leq (\sqrt{\rho(s)}-\sqrt{\rho(0)})/s \leq \sqrt{\rho(s)}/s.\]
    The function
    \[f_k(s)=\Big(1+\kappa+\sqrt{\rho(s_*+s_k+s)}\Big)^2\]
    is also increasing and concave, and by the above, its derivative at $s=0$ satisfies
    \[f_k'(0) \leq d_k/(s_*+s_k), \qquad d_k \equiv 2(1+\kappa)\sqrt{\rho(s_*+s_k)}+\rho(s_*+s_k).\]
    Thus
    \begin{equation}\label{eq:taukbound}
    (1+\tau_k)^2=f_k(s_{k+1}) \leq f_k(0)+f_k'(0) \cdot s_{k+1}
     \leq (1+\zeta_k)^2+d_ks_{k+1}/s_*.
    \end{equation}
    Applying this and (\ref{eq:tmpbound3}) to (\ref{eq:tmpbound2}), we get
   \begin{align*}
   \lambda_k s_{k+1} &\leq \left(1+(1+\tau_k)^2(1+\zeta_k)^2\right)\|\r_k\|_2^2+ 2\lambda_k s_*\\
   &\leq \left(1+(1+\zeta_k)^4+(1+\zeta_k)^2d_ks_{k+1}/s_*)\right)\|\r_k\|_2^2+2\lambda_k s_*.
   \end{align*}
   Rearranging gives
   \begin{equation}\label{eq:sbound}
   \Big(\lambda_k-(1+\zeta_k)^2d_k\|\r_k\|_2^2/s_*\Big)\cdot s_{k+1}
   \leq(1+(1+\zeta_k)^4)\cdot\|\r_k\|_2^2+2\lambda_k s_*.
   \end{equation}

   Second, applying the $(\kappa,\rho)$-cRIP condition for $\sqrt{\eta} \cdot \A$ again, we have for every $\u \in \R^p$
    \begin{align*}
        \Big|\u^\T(\eta\bP\A^\T\A\bP-\bP)\u\Big|
        &=\Big|\|\sqrt{\eta} \A\bP\u\|_2^2-\|\bP\u\|_2^2\Big|\\
        &\leq \max\Big(|1-(1-\tau_k)^2|,|1-(1+\tau_k)^2|\Big)\|\bP\u\|_2^2=(2\tau_k+\tau_k^2)\|\bP\u\|_2^2,
    \end{align*}
    So $\|\eta\bP\A^\T\A\bP-\bP\|_{\op} \leq 2\tau_k+\tau_k^2$.
    Then, as $\r_k=\bP\r_k$, we get from (\ref{eq:tmpbound}) that
    \[\Big(\|\r_{k+1}\|_2-(2\tau_k+\tau_k^2)\|\r_k\|_2\Big)_+^2
        \leq (2\tau_k+\tau_k^2)^2\|\r_k\|_2^2+\lambda_k(4s_*-2s_{k+1}).\]
    Taking the square-root and applying $\sqrt{x+y} \leq \sqrt{x}+\sqrt{y}$,
    \begin{align*}
        \|\r_{k+1}\|_2 \leq (4\tau_k+2\tau_k^2)\|\r_k\|_2
        +\sqrt{\lambda_k(4s_*-2s_{k+1})_+}
    \end{align*}
    Applying the definitions of $\tau_k$ and $t(\kappa)$,
    \[4\tau_k+2\tau_k^2 \leq 1-t(\kappa)+ 4(1+\kappa)\sqrt{\rho(s_*+s_k+s_{k+1})}+2\rho(s_*+s_k+s_{k+1}).\]
    Thus
    \begin{align}\label{eq:rbound}
    \|\r_{k+1}\|_2\leq\Big[1-t(\kappa)+4(1+\kappa)\sqrt{\rho(s_*+s_k+s_{k+1})}+2\rho(s_*+s_k+s_{k+1})\Big]\cdot\|\r_k\|_2+\sqrt{4\lambda_k s_*}.
    \end{align}
    
    We now claim by induction on $k$ that, if $\rho(s_*) \leq c_0$ for a sufficiently small constant $c_0>0$, then
        \begin{equation}\label{eq:induction}
            s_k\leq \frac{90}{t(\kappa)^2} s_*, \qquad
        \|\r_k\|_2 \leq \frac{4\sqrt{\lambda_k s_*}}{t(\kappa)}
        \end{equation}
    for every $k$. For $k=0$, these are satisfied as
    $s_0=0$ and $\lambda_0=\lambda_{\max} \geq \|\r_0\|_2^2=\|\x_*\|_2^2$.
    Assume inductively that these hold for $k$. Note that for any $t \geq 1$, nonnegativity and concavity yield $\rho(ts_*) \leq t\rho(s_*)$. In particular, assuming (\ref{eq:induction}) and applying $\kappa<\sqrt{3/2}-1$ and $\rho(s_*) \leq c_0$, we get for small enough $c_0$ that $(1+\zeta_k)^2<2$. Then applying (\ref{eq:induction}) to (\ref{eq:sbound}),
    we get for a constant $C \equiv C(\kappa)>0$ not depending on $c_0$ that
    \begin{align*}
    \left(1-C\sqrt{c_0}\right)\lambda_k s_{k+1}
    \leq \left(\frac{80}{t(\kappa)^2}+2\right)\lambda_k s_*.
    \end{align*}    
    Then for small enough $c_0$,
    \[s_{k+1} \leq \left(1-C\sqrt{c_0}\right)^{-1}\frac{82}{t(\kappa)^2}s_*<
    \frac{90}{t(\kappa)^2}s_*.\]
    Applying (\ref{eq:induction}) and this bound to (\ref{eq:rbound}),
    for sufficiently small $c_0$, we have
    \[\|\r_{k+1}\|_2 \leq \left(1-\frac{3}{4}t(\kappa)\right)\|\r_k\|_2
    +\sqrt{4\lambda_ks_*}
    \leq \left(\frac{4}{t(\kappa)}-1\right)\sqrt{\lambda_ks_*}.\]
    Applying $\sqrt{\lambda_k}=\sqrt{\lambda_{k+1}/\gamma}
    \leq \sqrt{\lambda_{k+1}}(1-t(\kappa)/4)^{-1}$, we obtain from this
    \[\|\r_{k+1}\|_2 \leq
    \frac{4\sqrt{\lambda_{k+1}s_*}}{t(\kappa)}.\]
    This completes the induction and establishes (\ref{eq:induction}) for every $k$.

    The bound (\ref{eq:recoverynonoise}) follows from (\ref{eq:induction}), the definition of $\r_k$, and $\lambda_k=\lambda_{\max}\gamma^k$.
    Since $\x_k,\x_* \in (\delta \Z)^p$, for $k$ large enough such that the right side of (\ref{eq:recoverynonoise}) is less than $\delta^2$, we must have $\x_k=\x_*$.
\end{proof}

We now extend this result to provide a robust recovery guarantee in the presence of measurement and discretization error. The proof is an extension of the above argument, which we defer to Appendix \ref{appendix:recovery}.

\begin{theorem}\label{thm:oracle}
    Suppose $\sqrt{\eta} \cdot \A$ satisfies $(\kappa,\rho)$-cRIP, where $\kappa \in [0,\sqrt{3/2}-1)$. Choose tuning parameters $\gamma,\lambda_{\max}$ as in Theorem \ref{thm:noiseless}.
    Then for some constants $C,C',c>0$ depending only on $\kappa$, the following holds: Let $\x \in (\delta\Z)^p$ be any vector satisfying
    \[\rho(s) \leq c, \qquad s \equiv \max(\|\nabla \x\|_0,1).\]
    Let $D$ be the maximum vertex degree of $G$, and define
    \[E(\x)=\left(1+\sqrt{D\rho(s)}\right)\cdot
    \left(\|\x-\x_*\|_2+\frac{\|\x-\x_*\|_1}{\sqrt{s}}\right)+\sqrt{\eta} \cdot \|\e\|_2.\]
    Suppose $\lambda_{\max} \geq CE(\x)^2/s \geq \lambda_{\min}$, and let $k_*$ be the last iterate of Algorithm \ref{alg:ITALE} where $\lambda_{k_*} \geq CE(\x)^2/s$. Then $\hat{\x} \equiv \x_{k_*}$ satisfies
    \[\|\hat{\x}-\x_*\|_2 \leq C'E(\x).\]
\end{theorem}

The quantity $E(\x)$ above is the combined measurement error and approximation
error of $\x_*$ by a discretized piecewise-constant signal $\x$. For any $\A$ scaled such that it satisfies $(\kappa,\rho)$-cRIP with $\eta=1$, and for $G$ with maximum degree $D \lesssim 1$, we get
\[\|\hat{\x}-\x_*\| \lesssim \|\x_*-\x\|_2+\frac{\|\x_*-\x\|_1}{\sqrt{s}}+\|\e\|_2.\]
This guarantee is similar to those for compressed sensing of sparse signals
in \cite{candesstable,needelltropp,blumensathdavies}.
If $\x_*$ has exact gradient-sparsity $\|\nabla \x_*\|_0 \leq s$, then also $\x
\in (\delta \Z)^p$ obtained by entrywise rounding to $\delta\Z$ satisfies
$\|\nabla \x\|_0 \leq s$. Hence choosing $\delta \ll \|\e\|_2/p$ further ensures
\[\|\hat{\x}-\x_*\| \lesssim \|\e\|_2\]
i.e.\ the discretization error is negligible in the above bound.
The required number of measurements is the same as in Theorem \ref{thm:noiseless} for the noiseless setting, which is $n \gtrsim s_*\log(1+|E|/s_*)$ for i.i.d.\ Gaussian designs. This is the claim (\ref{eq:heuristicguarantee}) stated in the introduction.

\section{Simulations}\label{sec:simulations}

We compare $\hat{\x}^\ITALE$ using the
$\ell_0$ edge cost (\ref{eq:l0cost}) with $\hat{\x}^\TV$ minimizing
the TV-regularized objective (\ref{eq:TVobjective}), for several signals on the
1-D and 2-D lattice graphs. We used software developed by
\cite{boykovkolmogorov} to implement the alpha-expansion sub-routine of
Algorithm \ref{alg:AlphaExpansion}. To minimize the TV-regularized objective
(\ref{eq:TVobjective}), we used the generalized lasso path algorithm from
\cite{rjtibshirani} in the 1-D examples and the FISTA algorithm from
\cite{beckteboulle} in the 2-D examples. All parameters were set as
described in Section \ref{sec:model} for ITALE.

\subsection{1-D changepoint signals}\label{sec:1D}

\begin{figure}
\begin{minipage}[t]{\textwidth}
\includegraphics[width=\textwidth]{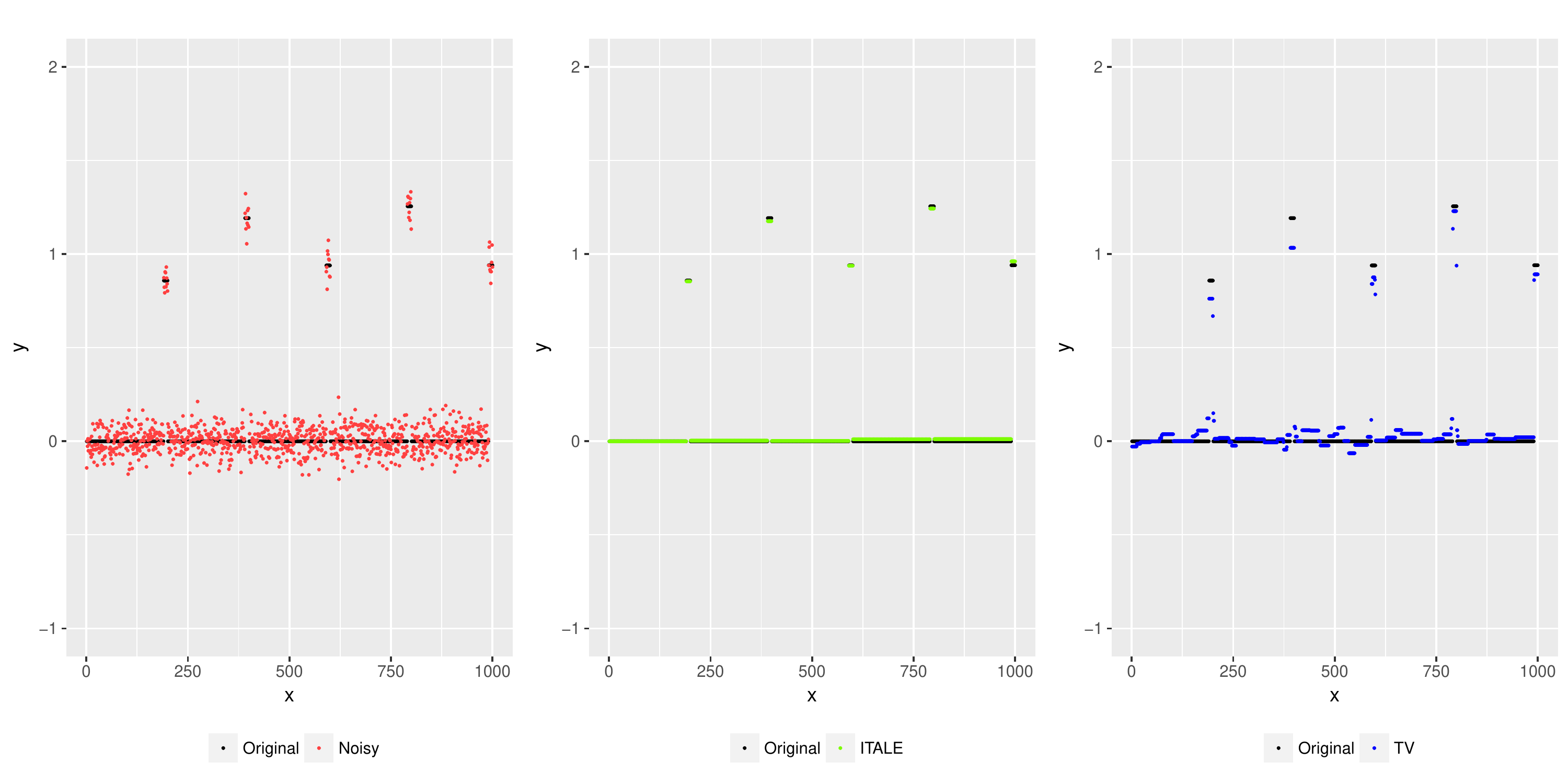}
\caption{Left: True spike signal $\x_*$ (black) and a depiction of $\x_*+\A^\T\e/n$ (red) under low noise $\sigma=1$ for i.i.d.\ measurements $A_{ij} \sim \N(0,1)$ with 15\% undersampling. Middle and right: True signal (black), $\hat{\x}^\ITALE$ (green), and $\hat{\x}^\TV$ (blue) for one simulation.}\label{fig:spike_150_1}
\end{minipage}
 
\begin{minipage}[t]{\textwidth}
\includegraphics[width=\textwidth]{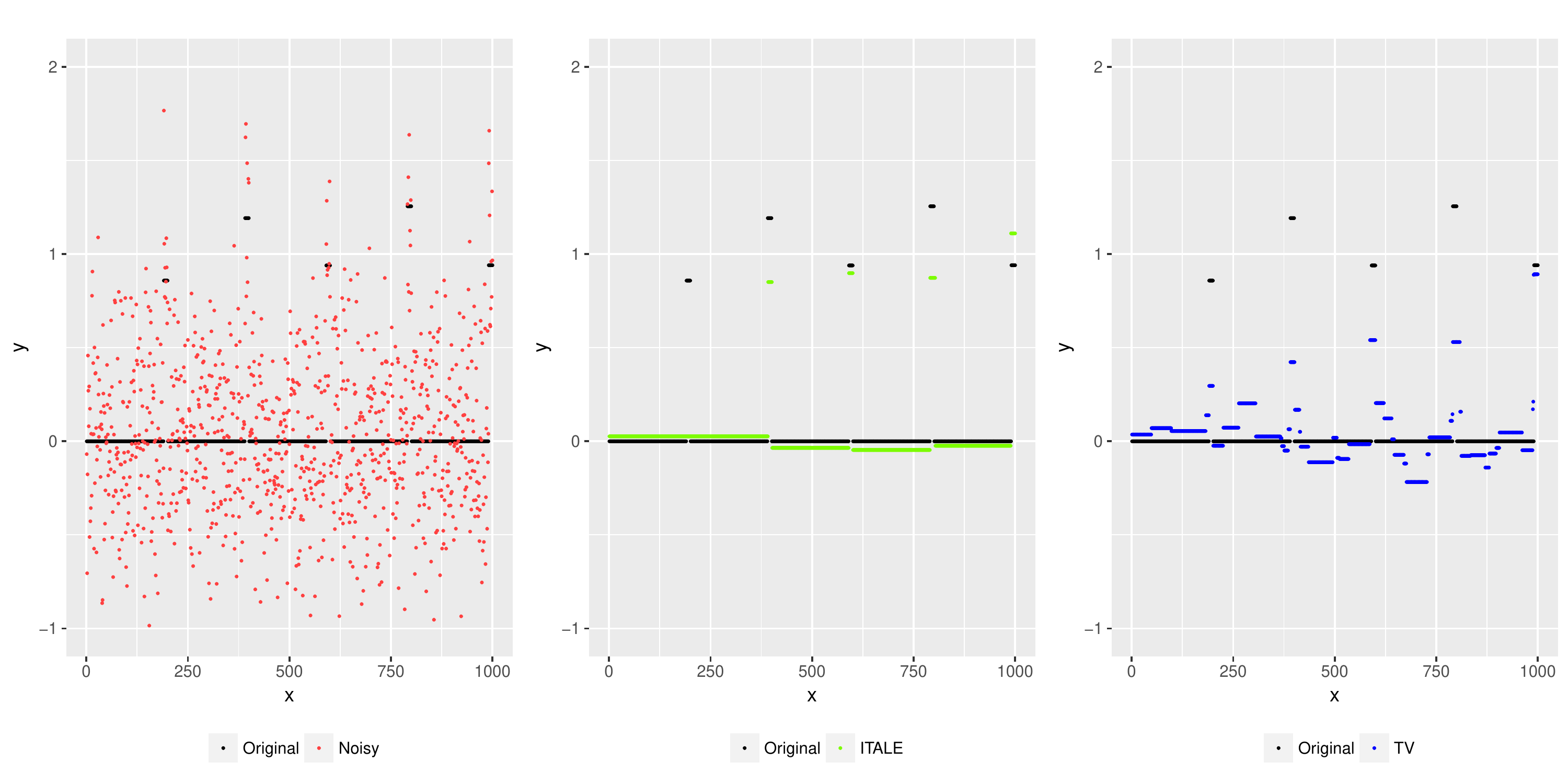}
\caption{Same setting as Figure \ref{fig:spike_150_1}, for noise level $\sigma=6$.}
\label{fig:spike_150_6}
\end{minipage}
\end{figure}

\begin{figure}
\begin{minipage}[t]{\textwidth}
\includegraphics[width=\textwidth]{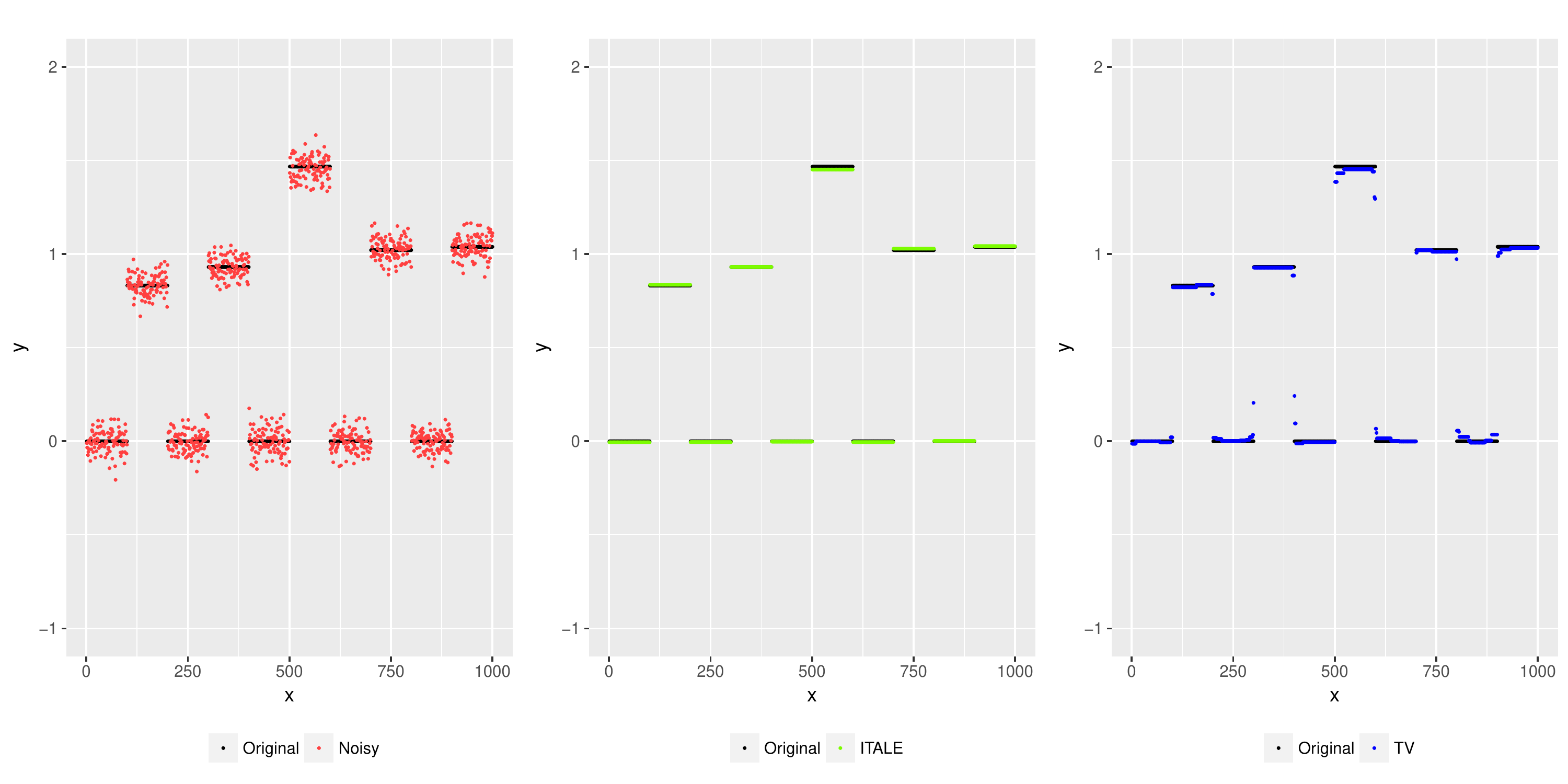}
\caption{Left: True wave signal $\x_*$ (black) and a depiction of
$\x_*+\A^\T\e/n$ (red) under low noise $\sigma=1$ for i.i.d.\ measurements
$A_{ij} \sim \N(0,1)$ with 15\% undersampling. Middle and right: True signal
(black), $\hat{\x}^\ITALE$ (green), and $\hat{\x}^\TV$ (blue) for one
simulation.}\label{fig:wave_150_1}
\end{minipage}
 
\begin{minipage}[t]{\textwidth}
\includegraphics[width=\textwidth]{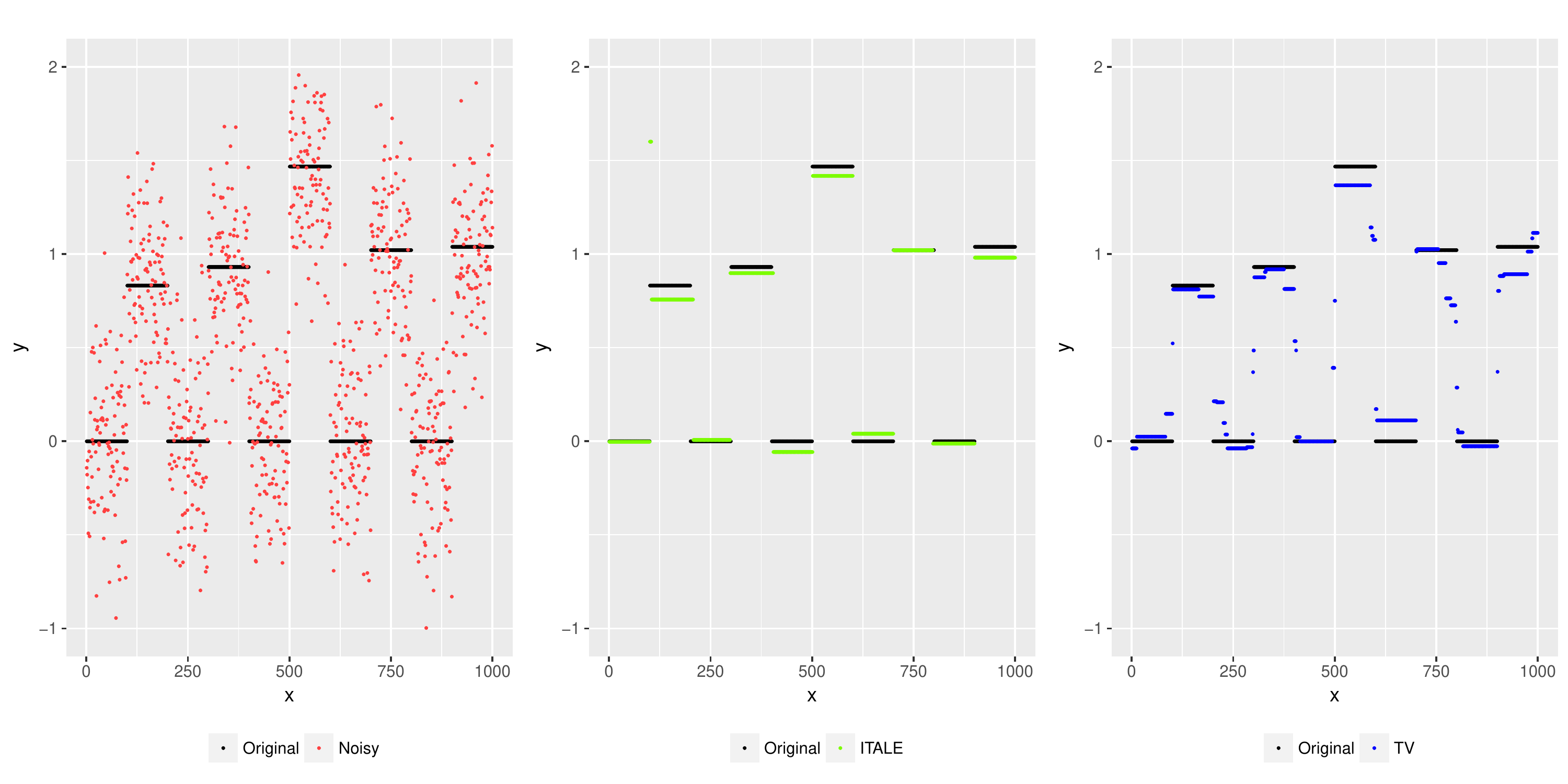}
\caption{Same setting as Figure \ref{fig:wave_150_1}, for noise level
$\sigma=6$.}\label{fig:wave_150_6}
\end{minipage}
\end{figure}

We tested ITALE on two simulated signals for the linear chain graph, with different changepoint
structures: the ``spike'' signal depicted in Figures \ref{fig:spike_150_1} and
\ref{fig:spike_150_6}, and the ``wave'' signal depicted in Figure
\ref{fig:wave_150_1} and \ref{fig:wave_150_6}. The two signals both have
$p=1000$ vertices with $s_*=9$ break points. The spike signal consists of
short segments of length 10 with elevated mean, while the breaks of the wave
signal are equally-spaced.

We sampled random Gaussian measurements $A_{ij} \overset{iid}{\sim} \N(0,1)$. The measurement error $\e$
was generated as Gaussian noise $e_k \overset{iid}{\sim} \N(0,\sigma^2)$. To
provide an intuitive understanding of the tested signal-to-noise,
we plot $\x_*+\A^\T\e/n$ in red in Figures \ref{fig:spike_150_1} to
\ref{fig:wave_150_6}, corresponding to two different tested noise levels. Recall
that ITALE denoises
$\a_{k+1}=\x_*+(\I-\A^\T\A/n)(\x_k-\x_*)+\A^\T\e/n$
in each iteration (corresponding to $\eta=1/n$
for this normalization of $\A$), so that
$\x_*+\A^\T\e/n$ represents the noisy signal in an ideal setting if $\x_k
\equiv \x_*$ is a perfect estimate from the preceding iteration.

Tables \ref{tab:spike} and \ref{tab:wave} display the root-mean-squared estimation errors
\[\text{RMSE} = \sqrt{\|\hat{\x}-\x_*\|_2^2/p},\]
for undersampling ratio $n/p$ from 10\% to 50\%, and a range of noise levels $\sigma$ that
yielded RMSE values between 0 and roughly 0.2. Each reported error value
is an average across 20 independent simulations. In these results,
the iterate $k$ in ITALE and penalty parameter $\lambda$ in TV were both
selected using 5-fold cross-validation. Best-achieved errors over all $k$ and
$\lambda$ are reported in Appendix \ref{appendix:bestachieved}, and suggest the
same qualitative conclusions. Standard deviations of the RMSE across simulations
are also reported in Appendix \ref{appendix:bestachieved}.

In the spike example, ITALE yielded lower RMSE in all of the above
settings of undersampling and signal-to-noise. Figures
\ref{fig:spike_150_1} and \ref{fig:spike_150_6} display one instance each of the 
resulting estimates $\hat{\x}^\ITALE$ and $\hat{\x}^\TV$ at 15\% undersampling,
illustrating some of their differences and typical features. Under optimal
tuning, $\hat{\x}^\TV$ returns an undersmoothed estimate even in a low-noise
setting where ITALE can often correctly estimate the changepoint locations. With
higher noise, ITALE begins to miss changepoints and oversmooth.

In the wave
example, with undersampling ranging between 15\% and 50\%, ITALE yielded lower
RMSE at most tested noise levels. Figures \ref{fig:wave_150_1} and
\ref{fig:wave_150_6} depict two instances of the recovered signals
at 15\% undersampling. For 10\% undersampling, the component
$(\I-\A^\T\A/n)(\x_k-\x_*)$ of the effective noise was sufficiently high such
that ITALE often did not estimate the true changepoint structure, and
TV usually outperformed ITALE in this case. The standard deviations of RMSE reported in
Appendix \ref{appendix:bestachieved} indicate that the ITALE estimates are
a bit more variable than the TV estimates in all tested settings, but
particularly so in this 10\% undersampling regime.

\begin{table}
\begin{tabular}{l|c||ccccccccc}
$n/p$ & & $\sigma=0$ & $\sigma=1$ & $\sigma=2$ & $\sigma=3$ & $\sigma=4$ & $\sigma=5$ & $\sigma=6$ & $\sigma=7$ \\
\hline
\multirow{2}{2em}{10\%} & ITALE & {\bf 0.000} & {\bf 0.014} & {\bf 0.060} & {\bf 0.090} & {\bf 0.144} & {\bf 0.173} & {\bf 0.199} & {\bf 0.216} \\
& TV & {\bf 0.000} & 0.047 & 0.092 & 0.129 & 0.160 & 0.189 & 0.213 & 0.228 \\
\hline
\multirow{2}{2em}{15\%} & ITALE & {\bf 0.000} & {\bf 0.009} & {\bf 0.023} & {\bf 0.049} & {\bf 0.076} & {\bf 0.104} & {\bf 0.133} & {\bf 0.153} \\
& TV & {\bf 0.000} & 0.030 & 0.060 & 0.088 & 0.114 & 0.136 & 0.158 & 0.175 \\
\hline
\multirow{2}{2em}{20\%} & ITALE & {\bf 0.000} & {\bf 0.007} & {\bf 0.015} & {\bf 0.032} & {\bf 0.056} & {\bf 0.076} & {\bf 0.099} & {\bf 0.123} \\
& TV & {\bf 0.000} & 0.022 & 0.045 & 0.067 & 0.089 & 0.109 & 0.128 & 0.146 \\
\hline
\multirow{2}{2em}{30\%} & ITALE & {\bf 0.000} & {\bf 0.006} & {\bf 0.012} & {\bf 0.021} & {\bf 0.031} & {\bf 0.049} & {\bf 0.065} & {\bf 0.079} \\
& TV & {\bf 0.000} & 0.017 & 0.035 & 0.052 & 0.070 & 0.087 & 0.104 & 0.120 \\
\hline
\multirow{2}{2em}{40\%} & ITALE & {\bf 0.000} & {\bf 0.005} & {\bf 0.010} & {\bf 0.015} & {\bf 0.025} & {\bf 0.041} & {\bf 0.051} & {\bf 0.063} \\
& TV & {\bf 0.000} & 0.014 & 0.028 & 0.043 & 0.057 & 0.071 & 0.085 & 0.098 \\
\hline
\multirow{2}{2em}{50\%} & ITALE & {\bf 0.000} & {\bf 0.005} & {\bf 0.010} & {\bf 0.015} & {\bf 0.023} & {\bf 0.033} & {\bf 0.040} & {\bf 0.051} \\
& TV & {\bf 0.000} & 0.013 & 0.026 & 0.038 & 0.051 & 0.064 & 0.075 & 0.088 \\
\hline
\end{tabular}

\caption{RMSE for the 1-D spike signal, averaged over 20 simulations.}\label{tab:spike}
\end{table}

\begin{table}
\begin{tabular}{l|c||ccccccccc}
$n/p$ & & $\sigma=0$ & $\sigma=1$ & $\sigma=2$ & $\sigma=3$ & $\sigma=4$ & $\sigma=5$ & $\sigma=6$ & $\sigma=7$ \\
\hline
\multirow{2}{2em}{10\%} & ITALE & 0.036 & 0.040 & 0.118 & 0.150 & 0.198 & 0.236 & 0.262 & 0.315 \\
& TV & {\bf 0.000} & {\bf 0.032} & {\bf 0.064} & {\bf 0.093} & {\bf 0.120} & {\bf 0.143} & {\bf 0.168} & {\bf 0.189} \\
\hline
\multirow{2}{2em}{15\%} & ITALE & {\bf 0.000} & {\bf 0.009} & {\bf 0.025} & {\bf 0.059} & 0.090 & 0.111 & 0.143 & 0.176 \\
& TV & {\bf 0.000} & 0.023 & 0.046 & 0.068 & {\bf 0.089} & {\bf 0.109} & {\bf 0.127} & {\bf 0.144} \\
\hline
\multirow{2}{2em}{20\%} & ITALE & {\bf 0.000} & {\bf 0.007} & {\bf 0.017} & {\bf 0.039} & {\bf 0.061} & {\bf 0.079} & {\bf 0.103} & {\bf 0.121} \\
& TV & {\bf 0.000} & 0.019 & 0.037 & 0.056 & 0.074 & 0.092 & 0.108 & 0.124 \\
\hline
\multirow{2}{2em}{30\%} & ITALE & {\bf 0.000} & {\bf 0.006} & {\bf 0.012} & {\bf 0.019} & {\bf 0.035} & {\bf 0.051} & {\bf 0.065} & {\bf 0.085} \\
& TV & {\bf 0.000} & 0.014 & 0.028 & 0.042 & 0.056 & 0.070 & 0.084 & 0.097 \\
\hline
\multirow{2}{2em}{40\%} & ITALE & {\bf 0.000} & {\bf 0.005} & {\bf 0.011} & {\bf 0.018} & {\bf 0.027} & {\bf 0.037} & {\bf 0.052} & {\bf 0.064} \\
& TV & {\bf 0.000} & 0.012 & 0.024 & 0.037 & 0.049 & 0.061 & 0.073 & 0.085 \\
\hline
\multirow{2}{2em}{50\%} & ITALE & {\bf 0.000} & {\bf 0.005} & {\bf 0.010} & {\bf 0.016} & {\bf 0.024} & {\bf 0.033} & {\bf 0.044} & {\bf 0.055} \\
& TV & {\bf 0.000} & 0.011 & 0.022 & 0.033 & 0.043 & 0.054 & 0.065 & 0.075 \\
\hline
\end{tabular}

\caption{RMSE for the 1-D wave signal, averaged over 20 simulations.}\label{tab:wave}
\end{table}

\subsection{2-D phantom images}\label{sec:2D}

\begin{figure}
\begin{minipage}{0.25\textwidth}
\includegraphics[width=0.9\textwidth]{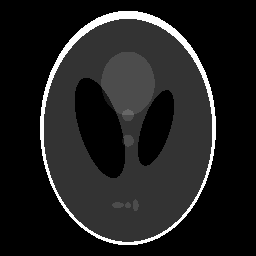}
\end{minipage}%
\begin{minipage}{0.5\textwidth}
\includegraphics[width=0.45\textwidth]{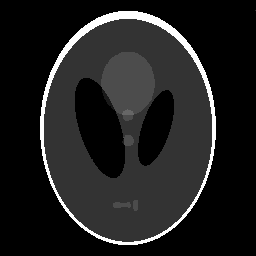}%
\includegraphics[width=0.45\textwidth]{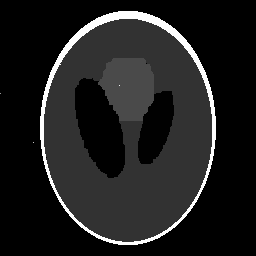}\\
\includegraphics[width=0.45\textwidth]{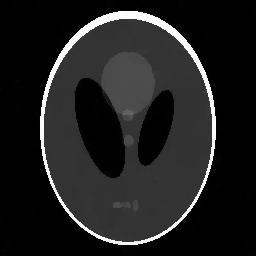}%
\includegraphics[width=0.45\textwidth]{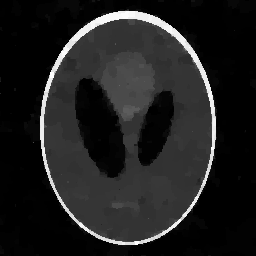}%
\end{minipage}
\caption{Left: Original Shepp-Logan phantom. Top row: $\hat{\x}^\ITALE$ from
15\% undersampled and reweighted Fourier measurements, in low noise ($\sigma=4$,
    left) and medium noise ($\sigma=16$, right) settings. Bottom row: $\hat{\x}^\TV$ for the same measurements.}\label{fig:shepplogan}
\end{figure}

\begin{figure}
\begin{minipage}{0.25\textwidth}
\includegraphics[width=0.8\textwidth]{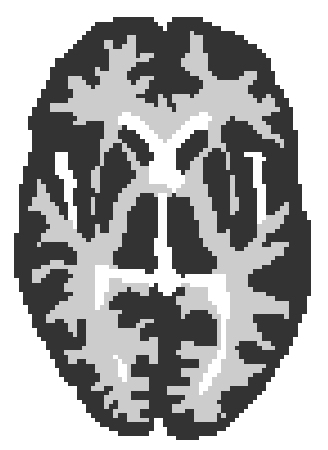}
\end{minipage}%
\begin{minipage}{0.5\textwidth}
\includegraphics[width=0.4\textwidth]{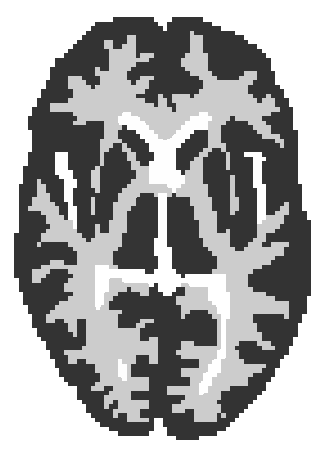}%
\includegraphics[width=0.4\textwidth]{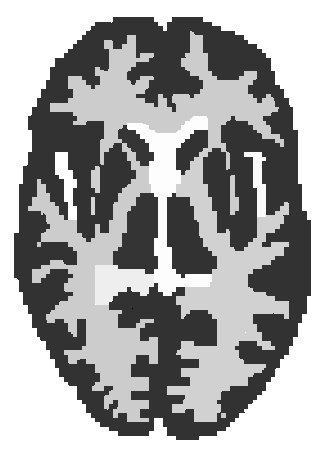}\\
\includegraphics[width=0.4\textwidth]{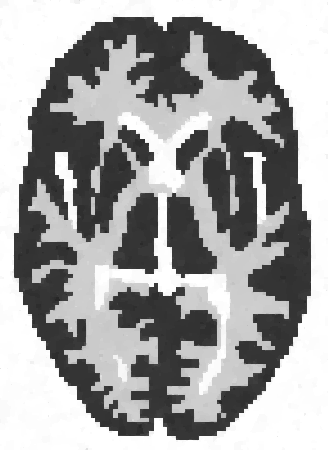}%
\includegraphics[width=0.4\textwidth]{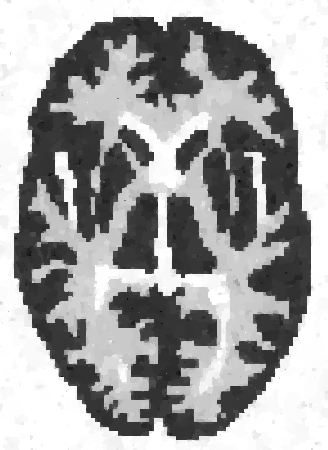}%
\end{minipage}
\caption{Left: Original brain phantom. Top row: $\hat{\x}^\ITALE$ from
20\% undersampled reweighted Fourier measurements, in low noise ($\sigma=16$, left) and medium noise ($\sigma=40$, right) settings. Bottom row: $\hat{\x}^\TV$ for the same measurements.}\label{fig:brain}
\end{figure}

Next, we tested ITALE on three 2-D image examples, corresponding to piecewise-constant digital phantom images of varying complexity: the Shepp-Logan digital phantom depicted in Figure \ref{fig:shepplogan}, a digital brain phantom from \cite{fesslerhero} depicted in Figure \ref{fig:brain}, and the XCAT chest slice from \cite{gongetal} as previously depicted in Figure \ref{fig:XCAT}.

\begin{table}
\begin{tabular}{l|c||cccccccc}
$n/p$ & & $\sigma=0$ & $\sigma=4$ & $\sigma=8$ & $\sigma=12$ & $\sigma=16$ & $\sigma=20$ & $\sigma=24$ & $\sigma=28$ \\
\hline
\multirow{2}{2em}{10\%} & ITALE & {\bf 0.001} & {\bf 0.007} & {\bf 0.012} & {\bf 0.018} & {\bf 0.023} & {\bf 0.039} & {\bf 0.047} & 0.072 \\
& TV & 0.005 & 0.011 & 0.021 & 0.031 & 0.040 & 0.051 & 0.057 & {\bf 0.062} \\
\hline
\multirow{2}{2em}{15\%} & ITALE & {\bf 0.000} & {\bf 0.003} & {\bf 0.011} & {\bf 0.014} & {\bf 0.017} & {\bf 0.027} & {\bf 0.035} & {\bf 0.044} \\
& TV & 0.001 & 0.008 & 0.016 & 0.026 & 0.030 & 0.038 & 0.049 & 0.053 \\
\hline
\multirow{2}{2em}{20\%} & ITALE & {\bf 0.000} & {\bf 0.003} & {\bf 0.008} & {\bf 0.012} & {\bf 0.017} & {\bf 0.024} & {\bf 0.027} & {\bf 0.036} \\
& TV & {\bf 0.000} & 0.007 & 0.014 & 0.021 & 0.027 & 0.032 & 0.039 & 0.045 \\
\hline
\multirow{2}{2em}{30\%} & ITALE & {\bf 0.000} & {\bf 0.002} & {\bf 0.005} & {\bf 0.011} & {\bf 0.013} & {\bf 0.015} & {\bf 0.018} & {\bf 0.026} \\
& TV & {\bf 0.000} & 0.006 & 0.012 & 0.017 & 0.023 & 0.027 & 0.032 & 0.036 \\
\hline
\multirow{2}{2em}{40\%} & ITALE & {\bf 0.000} & {\bf 0.001} & {\bf 0.005} & {\bf 0.010} & {\bf 0.011} & {\bf 0.013} & {\bf 0.017} & {\bf 0.018} \\
& TV & {\bf 0.000} & 0.005 & 0.010 & 0.015 & 0.019 & 0.023 & 0.028 & 0.032 \\
\hline
\multirow{2}{2em}{50\%} & ITALE & {\bf 0.000} & {\bf 0.001} & {\bf 0.004} & {\bf 0.008} & {\bf 0.011} & {\bf 0.013} & {\bf 0.013} & {\bf 0.015} \\
& TV & {\bf 0.000} & 0.005 & 0.009 & 0.013 & 0.017 & 0.021 & 0.025 & 0.029 \\
\hline
\end{tabular}

\caption{RMSE for the Shepp-Logan phantom}\label{tab:shepplogan}
\end{table}

\begin{table}
\begin{tabular}{l|c||cccccccc}
$n/p$ & & $\sigma=0$ & $\sigma=8$ & $\sigma=16$ & $\sigma=24$ & $\sigma=32$ & $\sigma=40$ & $\sigma=48$ & $\sigma=56$ \\
\hline
\multirow{2}{2em}{10\%} & ITALE & 0.003 & {\bf 0.002} & {\bf 0.010} & {\bf 0.028} & {\bf 0.043} & {\bf 0.060} & 0.080 & 0.096 \\
& TV & {\bf 0.002} & 0.014 & 0.028 & 0.042 & 0.054 & 0.064 & {\bf 0.078} & {\bf 0.085} \\
\hline
\multirow{2}{2em}{15\%} & ITALE & {\bf 0.000} & {\bf 0.001} & {\bf 0.007} & {\bf 0.019} & {\bf 0.030} & {\bf 0.044} & {\bf 0.059} & 0.075 \\
& TV & 0.001 & 0.011 & 0.022 & 0.032 & 0.043 & 0.053 & 0.063 & {\bf 0.074} \\
\hline
\multirow{2}{2em}{20\%} & ITALE & {\bf 0.000} & {\bf 0.001} & {\bf 0.005} & {\bf 0.012} & {\bf 0.025} & {\bf 0.036} & {\bf 0.047} & {\bf 0.056} \\
& TV & {\bf 0.000} & 0.009 & 0.019 & 0.028 & 0.038 & 0.046 & 0.056 & 0.061 \\
\hline
\multirow{2}{2em}{30\%} & ITALE & {\bf 0.000} & {\bf 0.001} & {\bf 0.002} & {\bf 0.007} & {\bf 0.014} & {\bf 0.025} & {\bf 0.033} & {\bf 0.044} \\
& TV & {\bf 0.000} & 0.007 & 0.016 & 0.023 & 0.030 & 0.038 & 0.046 & 0.053 \\
\hline
\multirow{2}{2em}{40\%} & ITALE & {\bf 0.000} & {\bf 0.000} & {\bf 0.002} & {\bf 0.006} & {\bf 0.010} & {\bf 0.021} & {\bf 0.027} & {\bf 0.037} \\
& TV & {\bf 0.000} & 0.007 & 0.013 & 0.019 & 0.026 & 0.032 & 0.039 & 0.043 \\
\hline
\multirow{2}{2em}{50\%} & ITALE & {\bf 0.000} & {\bf 0.001} & {\bf 0.002} & {\bf 0.005} & {\bf 0.007} & {\bf 0.016} & {\bf 0.022} & {\bf 0.029} \\
& TV & {\bf 0.000} & 0.006 & 0.011 & 0.018 & 0.023 & 0.029 & 0.036 & 0.040 \\
\hline
\end{tabular}

\caption{RMSE for the brain phantom}\label{tab:brain}
\end{table}

\begin{table}
\begin{tabular}{l|c||cccccccc}
$n/p$ & & $\sigma=0$ & $\sigma=4$ & $\sigma=8$ & $\sigma=12$ & $\sigma=16$ & $\sigma=20$ & $\sigma=24$ & $\sigma=28$ \\
\hline
\multirow{2}{2em}{10\%} & ITALE & 0.063 & 0.064 & 0.068 & 0.078 & 0.084 & 0.092 & 0.099 & 0.106 \\
& TV & {\bf 0.009} & {\bf 0.019} & {\bf 0.032} & {\bf 0.044} & {\bf 0.054} & {\bf 0.061} & {\bf 0.068} & {\bf 0.072} \\
\hline
\multirow{2}{2em}{15\%} & ITALE & {\bf 0.002} & {\bf 0.007} & 0.025 & 0.036 & 0.048 & 0.070 & 0.079 & 0.085 \\
& TV & 0.005 & 0.014 & {\bf 0.024} & {\bf 0.035} & {\bf 0.043} & {\bf 0.050} & {\bf 0.056} & {\bf 0.061} \\
\hline
\multirow{2}{2em}{20\%} & ITALE & {\bf 0.002} & {\bf 0.005} & {\bf 0.014} & {\bf 0.023} & {\bf 0.033} & 0.045 & 0.052 & 0.079 \\
& TV & {\bf 0.002} & 0.011 & 0.020 & 0.028 & 0.037 & {\bf 0.043} & {\bf 0.049} & {\bf 0.056} \\
\hline
\multirow{2}{2em}{30\%} & ITALE & {\bf 0.002} & {\bf 0.004} & {\bf 0.011} & {\bf 0.018} & {\bf 0.025} & {\bf 0.033} & {\bf 0.041} & 0.049 \\
& TV & {\bf 0.002} & 0.008 & 0.016 & 0.023 & 0.030 & 0.036 & 0.042 & {\bf 0.046} \\
\hline
\multirow{2}{2em}{40\%} & ITALE & 0.002 & {\bf 0.003} & {\bf 0.009} & {\bf 0.014} & {\bf 0.020} & {\bf 0.026} & {\bf 0.033} & {\bf 0.038} \\
& TV & {\bf 0.001} & 0.007 & 0.014 & 0.020 & 0.025 & 0.031 & 0.036 & 0.041 \\
\hline
\multirow{2}{2em}{50\%} & ITALE & 0.002 & {\bf 0.003} & {\bf 0.008} & {\bf 0.013} & {\bf 0.018} & {\bf 0.024} & {\bf 0.029} & {\bf 0.033} \\
& TV & {\bf 0.001} & 0.006 & 0.012 & 0.018 & 0.024 & 0.029 & 0.033 & 0.038 \\
\hline
\end{tabular}

\caption{RMSE for the XCAT chest slice phantom}\label{tab:XCAT}
\end{table}

Each image $\x_*$ was normalized to have pixel value in $[0,1]$.
We sampled random Fourier design matrices as specified in
(\ref{eq:weightedfourier}), fixing the constant $C_0=10$ in the weight
distribution (\ref{eq:nu}) for this design. This yielded the best recovery
across several tested values for both ITALE and TV. The measurement error $\e$
was generated as Gaussian noise $e_k \overset{iid}{\sim} \N(0,\sigma^2)$,
applied to the measurements $\F_{(i,j)}^*\x_*/\sqrt{\nu(i,j)}$ before the
$1/\sqrt{n}$ normalization. 
Tables \ref{tab:shepplogan}, \ref{tab:brain}, and \ref{tab:XCAT} display the
RMSE of the estimates $\hat{\x}^\ITALE$ and $\hat{\x}^\TV$
for a single simulation, with tuning parameters selected by 5-fold
cross-validation. Best-achieved errors are reported in Appendix
\ref{appendix:bestachieved}.

For the simpler Logan-Shepp and brain phantom images, which exhibit stronger
gradient-sparsity, ITALE yielded lower RMSE in nearly all tested undersampling
and signal-to-noise regimes. For the XCAT chest phantom, with undersampling
ranging between 15\% and 50\%, ITALE yielded lower RMSE at
a range of tested noise levels, and in particular for those settings of higher
signal-to-noise. With 10\% undersampling for the XCAT phantom, ITALE was not
able to recover some details of the XCAT image even with no measurement noise,
and RMSE was higher than TV at all tested noise levels. Results
of Appendix \ref{appendix:bestachieved} indicate that this is partially due to
sub-optimal selection of the tuning parameter using 5-fold cross-validation,
caused by the further reduction of undersampling from 10\% to 8\% in the size of
the training data in each fold.

Examples of recovered signals $\hat{\x}^\ITALE$ and $\hat{\x}^\TV$ are depicted
for the Shepp-Logan and brain phantoms in Figures \ref{fig:shepplogan} and
\ref{fig:brain}, at 15\% and 20\% undersampling for two low-noise and
medium-noise settings.
The qualitative comparisons are similar to those in the 1-D simulations, and to
those previously depicted for the XCAT chest slice in Figure \ref{fig:XCAT}: As
measurement noise increases, ITALE begins to lose finer
details, while TV begins to yield an undersmoothed and blotchy
image. These observations are also similar to previous comparisons that have
been made for algorithms oriented towards $\ell_0$ versus TV regularization
for direct measurements $\A=\I$, in \cite{xuetal,fanguan,kimgao}.

\section{Conclusion}
We have studied recovery of piecewise-constant signals over arbitrary graphs
from noisy linear measurements. We have proposed an iterative algorithm, ITALE, to minimize an $\ell_0$-edge-penalized least-squares objective. Under a
cut-restricted isometry property for the measurement design, we have established
global recovery guarantees for the estimated signal, in noisy and noiseless
settings.

In the field of compressed sensing, for signals exhibiting sparsity in an
orthonormal basis, $\ell_1$-regularization
\cite{donoho,candesstable,candesrobust} and discrete iterative algorithms
\cite{troppgilbert,needelltropp,blumensathdavies}
constitute two major approaches for signal recovery. It has been observed
that for recovering piecewise-constant signals, regularizing the signal gradient
in a sparse analysis framework can yield better empirical recovery than
regularizing signal coefficients in such a basis. Whereas
$\ell_1$-regularization extends naturally to the sparse analysis setting,
iterative algorithms have received less attention. By applying the
alpha-expansion idea for MAP estimation in discrete Markov random fields,
ITALE provides a computationally tractable approach for ``iterative
thresholding'' recovery of gradient-sparse signals, with provable recovery guarantees.

In contrast to sparse signal recovery over an orthonormal basis,
the comparison of $\ell_1$ versus $\ell_0$ regularization for gradient-based
sparsity is graph-dependent. Using an $\ell_0$-based approach, we establish
signal recovery guarantees on the 1-D and 2-D lattice graphs with numbers of
measurements optimal up to a constant factor, which were
not previously available for TV-regularization. This difference is closely
connected to slow and fast rates of convergence for lasso and best-subset
regression for correlated regression designs
\cite{buhlmannetal,zhangwainwrightjordan,dalalyanetal}. ITALE provides a
polynomial-time approach for $\ell_0$-regularization in a special graph-based
setting, and we believe it is an interesting question whether similar algorithmic
ideas may be applicable to other classes of sparse regression problems.

\appendix

\section{Proof of robust recovery guarantee}\label{appendix:recovery}

In this appendix, we prove Theorem \ref{thm:oracle} providing the estimation guarantee under approximate gradient-sparsity and discretization and measurement error.

\begin{lemma}\label{lem:error2}
Suppose $G$ has maximum vertex degree $D$, and $\A \in \R^{n\times p}$ satisfies $(\kappa,\rho)$-cRIP. Then for any $\u \in \R^p$ and $s \geq 1$,
\begin{align*}
\|\A\u\|_2 \leq \left(1+\kappa+\sqrt{D\rho(s)}\right)\cdot\left(\|\u\|_2+\frac{\|\u\|_1}{\sqrt{s}}\right).
\end{align*}
\end{lemma}

\begin{proof}
Let $T_1 \subseteq \{1,\ldots,p\}$ be the $s$ indices corresponding to the $s$ entries of $\u$ with largest magnitude (breaking ties arbitrarily). Let $T_2 \subseteq \{1,\ldots,p\} \setminus T_1$ be the $s$ indices corresponding to the next $s$ entries of $\u$ with largest magnitude, and define sequentially $T_3,T_4,\ldots,T_m$ for $m=\lceil p/s \rceil$ in this way. Denote by $\u_{T_i} \in \R^p$ the vector with $j^\text{th}$ entry equal to $u_j$ if $j \in T_i$, or 0 otherwise. Then $\|\nabla\u_{T_i}\|_0 \leq Ds$ for each $i$.
Applying the triangle inequality and cRIP condition for $\A$,
\[\|\A\u\|_2 \leq \sum_{i=1}^m \|\A\u_{T_i}\|_2
\leq \left(1+\kappa+\sqrt{\rho(Ds)}\right) \cdot \sum_{i=1}^m \|\u_{T_i}\|_2.\]
For $i \geq 2$, we have $\|\u_{T_{i+1}}\|_\infty \leq \|\u_{T_i}\|_1/s$ by construction, so
\begin{align*}
\|\u_{T_{i+1}}\|_2\leq \sqrt{s} \cdot\frac{\|\u_{T_i}\|_1}{s}=\frac{\|\u_{T_i}\|_1}{\sqrt{s}}.
\end{align*}
Applying this for $i \geq 2$, and the bound $\|\u_{T_1}\|_2 \leq \|\u\|_2$ for $i=1$,
\[\|\A\u\|_2 \leq \left(1+\kappa+\sqrt{\rho(Ds)}\right) \cdot \left(\|\u\|_2+\frac{\sum_{i=1}^{m-1} \|\u_{T_i}\|_1}{\sqrt{s}}\right) \leq \left(1+\kappa+\sqrt{\rho(Ds)}\right)\left(\|\u\|_2+\frac{\|\u\|_1}{\sqrt{s}}\right).\]
Finally, we have $(\rho(Ds)-\rho(0))/Ds \leq (\rho(s)-\rho(0))/s$ by the concavity of $\rho$, and hence $\rho(Ds) \leq D\rho(s)$ since $\rho(0) \geq 0$.
\end{proof}

\begin{proof}[Proof of Theorem \ref{thm:oracle}]
Write $\y=\A\x+\tilde{\e}$ where $\tilde\e=\A(\x_*-\x)+\e$. Denote
\[s=\max(\|\nabla\x\|_0,1), \qquad s_k=\|\nabla \x_k\|_0, \qquad \r_k=\x_k-\x.\]
As in the proof of Theorem \ref{thm:noiseless},
consider the partitions of $\{1,\ldots,p\}$ induced by the piecewise-constant
structures of $\x_k$, $\x_{k+1}$, and $\x$, let $\S$ be their common refinement,
and let $\bP$ be the orthogonal projection onto the subspace of signals taking constant value over each set in $\S$. Applying
\[\a_{k+1}=\x_k-\eta\A^\T(\A\x_k-\y)=\x_k-\eta \A^\T\A\r_k+\eta\A^\T\tilde{\e},\]
the same arguments as leading to (\ref{eq:tmpbound}) yield
    \begin{equation}\label{eq:ora_tmpbound}
        \left(\|\r_{k+1}\|_2-\|\r_k-\eta\bP\A^\T\A\r_k+\eta\bP\A^\T\tilde\e\|_2\right)_+^2
    \leq \|\r_k-\eta\bP\A^\T\A\r_k+\eta\bP\A^\T\tilde\e\|_2^2+\lambda_k(4s-2s_{k+1}).
    \end{equation}
Set $S_k=s+s_k+s_{k+1}$, $T_k=s+s_k$, and
    \[\tau_k=\kappa+\sqrt{\rho(S_k)},
    \quad \zeta_k=\kappa+\sqrt{\rho(T_k)},
    \quad d_k=2(1+\kappa)\sqrt{\rho(T_k)}+\rho(T_k).\]
    Then we obtain analogously to (\ref{eq:tmpbound2}) and (\ref{eq:sbound}) that
    \[\lambda_ks_{k+1} \leq \|\r_k\|_2^2+2\|\sqrt{\eta}\A\bP\|_{\op}^2 \cdot
    \|\sqrt{\eta}\A\r_k\|_2^2+2\|\eta\bP\A^\T\tilde{\e}\|_2^2+2\lambda_k s,\]
    and hence
\begin{equation}\label{eq:ora_sbound}
\left(\lambda_k-2(1+\zeta_k)^2d_k\|\r_k\|_2^2/s\right)\cdot s_{k+1} \leq (1+2(1+\zeta_k)^4)\cdot\|\r_k\|_2^2+2\|\eta \bP\A^\T\tilde\e\|_2^2+2\lambda_k s.
\end{equation}
Similarly, taking the square-root in (\ref{eq:ora_tmpbound}), we obtain analogously to (\ref{eq:rbound}) that
\begin{equation}\label{eq:ora_rbound}
    \|\r_{k+1}\|_2 \leq \Big[1-t(\kappa)+4\sqrt{\rho(S_k)}+2\rho(S_k)\Big]\cdot\|\r_k\|_2
    +2\|\eta\bP\A^\T\tilde{\e}\|_2+\sqrt{4\lambda_ks}.
\end{equation}

Recalling the bound $\|\sqrt{\eta} \A\bP\|_{\op} \leq 1+\tau_k$ from (\ref{eq:tmpbound3}), we have
\[\|\eta\bP\A^\T\tilde\e\|_2 \leq (1+\tau_k)(\|\sqrt{\eta}\A(\x_*-\x)\|_2+\|\sqrt{\eta} \cdot \e\|_2).\]
Bounding $\|\sqrt{\eta}\A(\x_*-\x)\|_2$ using the given cRIP condition and Lemma \ref{lem:error2} with the choice $s=\max(\|\nabla \x\|_0,1)$ as above, we get for a constant $c_1>0$ that
\[\|\eta\bP\A^\T\tilde\e\|_2 \leq (1+\tau_k)c_1E(\x).\]
Applying this and the bound $(1+\tau_k)^2 \leq (1+\zeta_k)^2+d_ks_{k+1}/s$ from (\ref{eq:taukbound}) to (\ref{eq:ora_sbound}), we get
\begin{equation}\label{eq:ora_sboundfinal}
\left(\lambda_k-2e_kd_k/s\right)
\cdot s_{k+1} \leq \|\r_k\|_2^2+2e_k(1+\zeta_k)^2+2\lambda_k s
\end{equation}
for the quantity
\[e_k=(1+\zeta_k)^2\|\r_k\|_2^2+c_1^2E(\x)^2.\]
Also, applying this to (\ref{eq:ora_rbound}), we get
\begin{equation}\label{eq:ora_rboundfinal}
    \|\r_{k+1}\|_2 \leq \Big[1-t(\kappa)+4\sqrt{\rho(S_k)}+2\rho(S_k)\Big]\cdot\|\r_k\|_2
    +2(1+\tau_k)c_1E(\x)+\sqrt{4\lambda_ks}.
\end{equation}

    We now claim by induction on $k$ that if $\rho(s) \leq c_0$ and
    $\lambda_k \geq C_0E(\x)^2/s$ for every $k \leq k_*$, where $C_0>0$ is sufficiently large and $c_0>0$ is sufficiently small, then for every $k \leq k_*$ we have
        \begin{equation}\label{eq:ora_induction}
            s_k\leq \frac{200}{t(\kappa)^2} s, \qquad
        \|\r_k\|_2 \leq \frac{4\sqrt{\lambda_k s}}{t(\kappa)}.
        \end{equation}
    For $k=0$, these are satisfied as
    $s_0=0$ and $\lambda_0=\lambda_{\max} \geq \|\r_0\|_2^2$.
    Assume inductively that these hold for $k$, where $k \leq k_*-1$. Then for small enough $c_0$, we have $(1+\zeta_k)^2<2$ and hence
    \[e_k \leq \frac{32\lambda_k s}{t(\kappa)^2}+c_1^2E(\x)^2.\]
    Also, $d_k \leq C\sqrt{c_0}$ for a constant $C \equiv C(\kappa)>0$
independent of $c_0$.
    Then for $C_0$ large enough and $c_0$ small enough, we obtain from (\ref{eq:ora_sboundfinal})
    and the condition $\lambda_k \geq C_0E(\x)^2/s$ that
\[\frac{3}{4}\lambda_k s_{k+1} \leq \|\r_k\|_2^2+4e_k+2\lambda_k s
\leq \frac{146\lambda_k s}{t(\kappa)^2}+4c_1^2E(\x)^2
    <\frac{150\lambda_k s}{t(\kappa)^2}.\]
    This gives the bound
    \[s_{k+1} \leq \frac{200}{t(\kappa)^2}s.\]

    Then applying (\ref{eq:ora_induction}) and this bound to (\ref{eq:ora_rboundfinal}),
    again for $\rho(s) \leq c_0$ and $\lambda_k s \geq C_0E(\x)^2$ with $C_0$ sufficiently large and $c_0$ sufficiently small, we get
    \[\|\r_{k+1}\|_2 \leq \left(1-\frac{4}{5}t(\kappa)\right)\|\r_k\|_2
    +6c_1E(\x)+\sqrt{4\lambda_ks}<\left(\frac{4}{t(\kappa)}-1\right)\sqrt{\lambda_ks}.\]
    Applying $\sqrt{\lambda_k}=\sqrt{\lambda_{k+1}/\gamma}
    \leq \sqrt{\lambda_{k+1}}(1-t(\kappa)/4)^{-1}$, we obtain
    \[\|\r_{k+1}\|_2 \leq
    \frac{4\sqrt{\lambda_{k+1}s}}{t(\kappa)},\]
    completing the induction. This establishes (\ref{eq:ora_induction}) for every $k \leq k_*$, provided $\lambda_{k_*} \geq C_0E(\x)^2/s$. In particular, at the iterate $k_*$, we have $\lambda_{k_*}s \asymp E(\x)^2$ and hence $\|\r_{k_*}\|_2 \lesssim E(\x)$.
\end{proof}

\section{Proofs of cut-restricted isometry property}\label{appendix:RIP}

In this appendix, we prove Propositions \ref{prop:subgaussian} and \ref{prop:2DFourier} establishing cRIP for the sub-Gaussian and weighted 2D-Fourier designs.

\begin{proof}[Proof of Proposition \ref{prop:subgaussian}]
First fix $s \in \{1,\ldots,|E|\}$.
For each partition $\S$ of $V=\{1,\ldots,p\}$ with $|\partial \S|=s$,
let $\bP_\S:\R^p \to K_\S$ be the associated orthogonal projection onto the subspace $K_\S$ of signals which are constant on each set in $\S$. Note that the dimension of $K_\S$ is the number of sets in $\S$, which is at most $s+1$ because $G$ is a connected graph. Write $\bP_\S=\Q_\S\Q_\S^\T$, where $\Q_\S$ has orthonormal columns spanning $K_\S$. Then $\A\Q_\S$ still has independent rows $\a_i^\T\Q_\S/\sqrt{n}$, where $\|\a_i^\T\Q_\S\|_{\psi_2} \leq K$ and $\operatorname{Cov}[\a_i^\T\Q_\S]=\Q_\S^\T\bSigma\Q_\S$. Applying \cite[Eq.\ (5.25)]{vershynin} to $\A\Q_\S$, for any $t>0$ and some constants $C,c>0$ depending only on $K$,
\begin{align*}
\P\left[\|\Q_\S^\T\A^\T\A\Q_\S-\Q_\S^\T\bSigma\Q_\S\|_{\op} \geq \max(\delta,\delta^2)\right]\leq 2e^{-ct^2},
\qquad \delta \equiv \frac{C\sqrt{s}+t}{\sqrt{n}}.
\end{align*}
Let $g(s)=s\log(1+|E|/s)$, and note that there are at most $\binom{|E|}{s} \leq e^{g(s)}$ partitions $\S$ where $|\partial \S|=s$. Taking a union bound over $\S$, and noting
that any $\u$ with $\|\nabla \u\|_0=s$ may be represented as $\u=\Q_{\S}\v$ for some such $\S$, this yields
\[\P\left[\sup_{\u \in \R^p:\|\u\|_2=1,\|\nabla \u\|_0=s} |\u^\T\A^\T\A\u-\u^\T\bSigma\u| \geq \max(\delta,\delta^2)\right] \leq 2e^{g(s)-ct^2}.\]
When $\|\u\|_2=1$ and $|\u^\T\A^\T\A\u-\u^\T\bSigma\u| \leq \max(\delta,\delta^2)$, we have
\[\|\A\u\|_2 \leq \sqrt{\u^\T\bSigma\u+\max(\delta,\delta^2)}
\leq \sqrt{(1+\kappa)^2+\max(\delta,\delta^2)}
\leq 1+\kappa+\delta.\]
We also have
\[\|\A\u\|_2 \geq \sqrt{(\u^\T\bSigma\u-\max(\delta,\delta^2))_+}
\geq \sqrt{((1-\kappa)^2-\max(\delta,\delta^2))_+}
\geq 1-\kappa-\frac{\delta}{1-\kappa},\]
where the last inequality is trivial for $\delta \geq (1-\kappa)^2$ and may be
checked for $\delta \leq (1-\kappa)^2$ by squaring both sides and applying
$\max(\delta,\delta^2)=\delta$ in this case. Then, for any $k$ and some
constants $C_0,C_1>0$ depending on $k$, setting
$t=\sqrt{C_0g(s)}$ and applying $g(s) \geq g(1)=\log (1+|E|)$, we get
\[\P\left[\;\sup_{\u \in \R^p:\|\u\|_2=1,\|\nabla \u\|_0=s} \Big|\|\A\u\|_2-1\Big| \leq \kappa+\sqrt{\frac{C_1g(s)}{n}}\;\right]
\leq |E|^{-k-1}.\]
Taking a union bound over $s=1,\dots,|E|$ and applying scale invariance of the cRIP condition to $\|\u\|_2$ concludes the proof.
\end{proof}

Next, we establish Proposition \ref{prop:2DFourier} on the Fourier design.

\begin{lemma}\label{lemma:ellinftybound}
    Let $p=N_1N_2$, let $S \times T \subset \{1,\ldots,N_1\} \times \{1,\ldots,N_2\}$ be any connected rectangle, and let $\F \in \C^{p \times p}$ be the 2-D discrete
    Fourier matrix defined in Section \ref{sec:RIP}. Then for any $(i,j) \in \{1,\ldots,N_1\} \times \{1,\ldots,N_2\}$,
    \[\left|\sum_{(i',j') \in S \times T}\F_{(i,j),(i',j')}\right|
    \leq \sqrt{\frac{|S|}{1+\min(i-1,N_1-i+1)} \cdot \frac{|T|}
    {1+\min(j-1,N_2-j+1)}}.\]
\end{lemma}
\begin{proof}
    Since
    \[\left|\sum_{(i',j') \in S \times T}\F_{(i,j),(i',j')}\right|
    =\left|\sum_{i' \in S} \F^1_{i,i'}
    \cdot \sum_{j' \in T} \F^2_{j,j'}\right|\]
    for the 1-D Fourier matrices $\F^1 \in \C^{N_1 \times N_1}$ and $\F^2 \in \C^{N_2 \times N_2}$, it suffices to show
    \[\left|\sum_{k \in S} \F_{ik}^1\right| \leq
    \sqrt{\frac{|S|}{1+\min(i-1,N_1-i+1)}}.\]
    For this, denote the elements of $S$ as $\{k_1+1,\ldots,k_1+|S|\}$, and write
    \[\left|\sum_{k \in S} \F^1_{ik}\right|
    =\left|\frac{1}{\sqrt{N_1}}
    \sum_{t=0}^{|S|-1} e^{2\pi \i \cdot \frac{(i-1)k_1}{N_1}}
    \cdot e^{2\pi \i \cdot \frac{(i-1)t}{N_1}}\right|
    =\frac{1}{\sqrt{N_1}}\left|
    \sum_{t=0}^{|S|-1} e^{2\pi \i \cdot \frac{(i-1)t}{N_1}}\right|.\]
    This is at most $|S|/\sqrt{N_1}$, which implies the bound for $i=1$. For $i \geq 2$, apply further
    \[|1-e^{2\pi \i t}| \geq 4\min(t,1-t)\]
    for $t \in [0,1]$. Then summing the geometric series, we also have
    \begin{align*}
        \left|\sum_{k \in S} \F_{ik}^1\right|
        &=\frac{1}{\sqrt{N_1}}\left|1-e^{2\pi \i \frac{(i-1)|S|}{N_1}}\right|
    \cdot \left|1-e^{2\pi \i \frac{(i-1)}{N_1}}\right|^{-1}\\
        &\leq \frac{\sqrt{N_1}}{2\min(i-1,N_1-i+1)}
    \leq \frac{\sqrt{N_1}}{1+\min(i-1,N_1-i+1)}.
    \end{align*}
    The result follows from combining this with the previous upper bound bound $|S|/\sqrt{N_1}$, using $\min(a,b) \leq \sqrt{ab}$.
\end{proof}

\begin{lemma}\label{lemma:Haardecay}
    Let $p=N_1N_2$, where $N_1,N_2$ are powers of 2 and $1/K \leq N_1/N_2 \leq K$ for a constant $K>0$. Let $G$ be the 2-D lattice graph of size $N_1 \times N_2$.
    For $\x \in \C^p$, let $|c_{(1)}(\x)| \geq \ldots \geq |c_{(p)}(\x)|$ be 
    the ordered magnitudes of the coefficients of $\x$ in the bivariate Haar
    wavelet basis. If $\x$ is centered to have mean entry 0, then for a
    constant $C \equiv C(K)>0$ and each $k=1,\ldots,p$,
    \[|c_{(k)}(\x)| \leq C \cdot \frac{\|\nabla \x\|_1}{k}\]
    where $\nabla$ is the discrete gradient operator on $G$.
\end{lemma}
\begin{proof}
    See \cite[Proposition 8]{needellward} for the case $N_1=N_2$. For $N_1<N_2$, we may apply this result to the ``stretched'' image where each original vertex value is copied to $N_2/N_1$ consecutive values in a vertical strip. This stretching changes $\|\nabla \x\|_1$ and each original bivariate Haar wavelet coefficient by at most a constant factor, and introduces $N_2^2-N_1N_2$ new Haar wavelet coefficients which are identically 0. Thus the result still holds in this case, and similarly for $N_1>N_2$.
\end{proof}

\begin{proof}[Proof of Proposition \ref{prop:2DFourier}]
    We follow closely the ideas of \cite[Theorem 3.3]{rudelsonvershynin}.

    For each partition $\S=(S_1,\ldots,S_k)$ of $G$ into $k$ connected pieces,
    let $K_\S \subset \C^p$ be the $k$-dimensional subspace of
    vectors which take a constant value over each set of $\S$.
    For each sparsity level $s \geq 1$, define
    \[K_s=\bigcup_{\S:|\partial \S| \leq s} \{\x \in K_\S:\|\x\|_2 \leq 1\},
    \qquad \kappa_s=\sup_{\x \in K_s} |\x^*(\A^*\A-\I)\x|.\]
    It suffices to show, with the stated probability and form of $\rho$, that
    \[\kappa_s \leq 2\sqrt{\rho(s)}+\rho(s)\]
    holds simultaneously for all $s=1,\ldots,|E|$.

    We first control $\E[\kappa_s]$ using a metric entropy argument:
    Letting $\A_r^*$ be row $r$ of $\A$,
    \[n \cdot \E[\A_r\A_r^*]
    =\E\left[\frac{\F_{(i_r,j_r)}\F_{(i_r,j_r)}^*}{\nu(i_r,j_r)}\right]
    =\sum_{(i,j)} \F_{(i,j)}\F_{(i,j)}^*=\I.\]
    So
    \[\kappa_s=\sup_{\x \in K_s} \cdot \left|\sum_{r=1}^n \left(|\A_r^*\x|^2
    -\E|\A_r^*\x|^2\right)\right|.\]
    Applying Gaussian symmetrization,
    \[\E[\kappa_s] \leq C\,\E \sup_{\x \in K_s}
    \left|\sum_{r=1}^n g_r |\A_r^*\x|^2\right|\]
    for a constant $C>0$ and $g_1,\ldots,g_n \overset{iid}{\sim} \N(0,1)$
    independent of $\A$.
    
    Condition on $\A$, and define by $E(\A)$ the right side above with the
    expectation taken only over $g_1,\ldots,g_n$. Introducing the pseudo-metric
    \[d(\x,\y)=\sqrt{\sum_{r=1}^n (|\A_r^*\x|^2-|\A_r^*\y|^2)^2},\]
    Dudley's inequality yields
    \[E(\A) \leq C\int_0^\infty \sqrt{\log N(K_s,d,u)}\,du\]
    where $N(K_s,d,u)$ is the covering number of $K_s$ by balls of radius
    $u$ in the metric $d$. For $\x,\y \in K_s$,
    \begin{align*}
        d(\x,\y) &\leq \sqrt{\sum_{r=1}^n
        |\A_r^*\x+\A_r^*\y|^2|\A_r^*\x-\A_r^*\y|^2}\\
        &\leq \sqrt{2\sup_{\z \in K_s} \sum_{r=1}^n |\A_r^*\z|^2} \cdot
        \max_{r=1}^n |\A_r^*\x-\A_r^*\y|=R(\A) \cdot \|\x-\y\|_\A,
    \end{align*}
    where
    \[R(\A)^2=\sup_{\z \in K_s} \z^*\A^*\A\z, \qquad
    \|\x\|_\A=\sqrt{2} \cdot \max_{r=1}^n |\A_r^*\x|.\]
    Applying this bound and a change-of-variables $v=u/R(\A)$,
    \begin{equation}\label{eq:dudleyintegral}
        E(\A) \leq CR(\A) \int_0^\infty \sqrt{\log N(K_s,\|\cdot\|_\A,v)}\,dv.
    \end{equation}

    The pseudo-norm $\|\cdot\|_\A$ has the following property:
    For any $\x \in K_s$,
    \begin{equation}\label{eq:TVbound}
        \|\nabla \x\|_1 \leq \sqrt{s}\|\nabla \x\|_2 \leq \sqrt{8s}\|\x\|_2
        \leq \sqrt{8s}
    \end{equation}
    where the middle inequality applies $(x-y)^2 \leq 2x^2+2y^2$ and the fact
    that the maximal vertex degree in $G$ is 4.
    Let $\v_1,\ldots,\v_p$ be the bivariate Haar wavelet basis, and write the
    orthogonal decomposition $\x=\sum_k c_k\v_k$.
    Then, as $\|\x\|_2 \leq 1$, $\|\nabla \x\|_1 \leq \sqrt{8s}$,
    and $\sum_{k=1}^p 1/k \leq C\log p$,
    Lemma \ref{lemma:Haardecay} implies
    \begin{equation}\label{eq:Haarl1}
        \sum_{k=1}^p |c_k| \leq C\sqrt{s}\log p.
    \end{equation}
    Each Haar vector $\v_k$ is supported on a number
    $\alpha \in \{1,2,4\}$ of rectangular pieces of some size $|S| \times |T|$,
    with a constant value $\pm 1/\sqrt{\alpha|ST|}$ on each piece. Then
    Lemma \ref{lemma:ellinftybound} implies for each $(i,j)$
    \[|\F_{(i,j)}^* \v_k| \leq C\sqrt{\frac{1}{1+\min(i-1,N_1-i+1)} \cdot
    \frac{1}{1+\min(j-1,N_2-j+1)}}.\]
    From the definition of $\nu$ and the bound $\sum_{k=1}^p 1/k \leq C\log p$,
    \[\nu(j) \geq \frac{c}{(\log p)^2}\cdot
    \frac{1}{C_0+\min(i-1,N_1-i+1)} \cdot \frac{1}{C_0+\min(j-1,N_2-j+1)}\]
    for a constant $c>0$. Then from the definitions of $\A$ and $\|\cdot\|_\A$,
    the bound (\ref{eq:Haarl1}), and the condition $1/K<N_1/N_2<K$, we obtain
    \begin{equation}\label{eq:Anormbound}
        \|\v_k\|_\A \leq C(\log p)/\sqrt{n}, \qquad
        \|\x\|_\A \leq B \equiv C(\log p)^2\sqrt{s/n}.
    \end{equation}

    As in \cite[Theorem 3.3]{rudelsonvershynin}, we bound the covering
    number $N(K_s,\|\cdot\|_\A,v)$ in two ways: First, fix any $\x \in K_s$ and
    write now its Haar decomposition as
    \[\x=\sum_{k=1}^p (a_k+\i b_k)\v_k\]
    where $a_k,b_k \in \R$. Then for some universal constant $L>0$, we obtain
    from (\ref{eq:Haarl1})
    \[\sum_{k=1}^p |a_k|+|b_k| \leq L\sqrt{s}\log p.\]
    Applying Maurey's argument, define a
    discrete distribution over a random vector $\z \in \C^p$ by
    \begin{align*}
        \P\Big[\z=L\sqrt{s}\log p \cdot
        \sign(a_k)\v_k\Big]&=\frac{|a_k|}{L\sqrt{s}\log p},\\
        \P\Big[\z=L\sqrt{s}\log p \cdot
        \sign(b_k)\i \v_k\Big]&=\frac{|b_k|}{L\sqrt{s}\log p},\\
        \P\Big[\z=0\Big]&=1-\sum_{k=1}^p \frac{|a_k|+|b_k|}{L\sqrt{s}
        \log p}.
    \end{align*}
    Then by construction, $\E[\z]=\x$. Letting $\z_1,\ldots,\z_m$ be
    independent copies of $\z$, for a value $m$ to be chosen later,
    Gaussian symmetrization yields (with all expectations conditional on $\A$)
    \begin{equation}\label{eq:Maurey}
        \E\left\|\x-\frac{1}{m}\sum_{j=1}^m \z_j\right\|_\A
        =\E \max_{r=1}^n \left|\frac{1}{m}\sum_{j=1}^m \A_r^*\z_j-
        \E \A_r^*\z_j\right|
    \leq \frac{C}{m}\E \max_{r=1}^n \left|\sum_{j=1}^m g_j \A_r^*\z_j\right|
    \end{equation}
    for $g_1,\ldots,g_m \overset{iid}{\sim} \N(0,1)$.
    The bound (\ref{eq:Anormbound}) yields for every $r$
    \[\sum_{j=1}^m |\A_r^*\z_j|^2 \leq \frac{Csm(\log p)^4}{n}.\]
    Applying this to (\ref{eq:Maurey}) with a Gaussian tail bound and union
    bound,
    \[\E\left\|\x-\frac{1}{m}\sum_{j=1}^m \z_j\right\|_\A
    \leq C\sqrt{\log n}\cdot (\log p)^2\sqrt{\frac{s}{mn}}.\]
    For any $v>0$, choosing $m=C(\log n)(\log p)^4 s/(nv^2)$
    ensures this bound is at most $v$. Then by the probabilistic method,
    $\x$ belongs to the $\|\cdot\|_\A$-ball of radius
    $v$ around some vector of the form $m^{-1} \sum_{j=1}^m \z_j$. The
    support of the distribution of $\z_j$ has cardinality at most $2p+1$, and
    this support is the same for all $\x \in K_s$. Then there are
    at most $(2p+1)^m$ such vectors, so we obtain
    \begin{equation}\label{eq:covering1}
        \sqrt{\log N(K_s,\|\cdot\|_\A,v)}
        \leq \sqrt{m \log (2p+1)}
        \leq C\sqrt{s(\log n)(\log p)^5/n} \cdot 1/v.
    \end{equation}

    We obtain a second covering bound by a union bound over $\S$:
    For any $\S=(S_1,\ldots,S_k)$ with $|\partial \S| \leq s$, note that
    $k \leq s+1$. Define
    $\U_\S \in \R^{p \times k}$ such that its $i$th column is
    $\e_{S_i}/\sqrt{|S_i|}$ where $\e_{S_i} \in \{0,1\}^p$
    is the indicator of $S_i$.
    Then $\U_\S \U_\S^*$ is the projection onto $K_\S$, and
    \[K_\S=\{\U_\S \y:\|\y\|_2 \leq 1\}.\]
    As $\max \{\|\U_\S \y\|_\A:\,\y \in \C^k,\|\y\|_2 \leq 1\} \leq B$
    by (\ref{eq:Anormbound}), a standard volume argument yields for $v \leq B$
    \[N(K_\S,\|\cdot\|_\A,v) \leq (CB/v)^k.\]
    The number of partitions $\S$ with $|\partial \S| \leq s$ is at most
    $\sum_{j=0}^s \binom{|E|}{j} \leq (Cp)^{s+1}$.
    Applying $k \leq s+1$ and summing over $\S$,
    \begin{equation}\label{eq:covering2}
        \sqrt{\log N(K_s,\|\cdot\|_\A,v)}
        \leq C\sqrt{s\log (CBp/v)}.
    \end{equation}

    Returning to the entropy integral in (\ref{eq:dudleyintegral}),
    note that (\ref{eq:Anormbound}) implies $N(K_s,\|\cdot\|_\A,v)=1$
    for $v>B$, so the integral may be restricted to $v \in [0,B]$.
    Setting $t=1/\sqrt{n}$, applying (\ref{eq:covering2}) for $v \in [0,t]$,
    and also applying Cauchy-Schwarz and $\log (B/t) \leq C\log p$, we get
    \begin{align*}
        \int_0^t \sqrt{\log N(K_s,\|\cdot\|_\A,v)}\,dv
        &\leq \sqrt{t} \cdot \sqrt{\int_0^t \log N(K_s,\|\cdot\|_\A,v)dv}\\
        &\leq Ct \sqrt{s\left(1+\log \frac{CBp}{t}\right)}
        \leq C\sqrt{\frac{s\log p}{n}}.
    \end{align*}
    Applying (\ref{eq:covering1}) for $v \in [t,B]$, we get
    \begin{align*}
        \int_t^B \sqrt{\log N(K_s,\|\cdot\|_\A,v)}\,dv
        &\leq C\sqrt{s(\log n)(\log p)^7/n}.
    \end{align*}
    Applying these bounds to (\ref{eq:dudleyintegral}) gives
    \[E(\A) \leq C\sqrt{s(\log n)(\log p)^7/n} \cdot R(\A).\]
    Taking now the expectation over $\A$ and applying Cauchy-Schwarz
    and the triangle inequality,
    \begin{align*}
        \E[\kappa_s] \leq \E[E(\A)]
        &\leq C\sqrt{s(\log n)(\log p)^7/n}\sqrt{\E[R(\A)^2]}\\
        &\leq C\sqrt{s(\log n)(\log p)^7/n}\sqrt{\E[\kappa_s]+1}.
    \end{align*}
    This yields
    \[\E[\kappa_s] \leq E(p,n,s)
    \equiv C \max\left(\sqrt{s(\log n)(\log p)^7/n},\;
    s(\log n)(\log p)^7/n\right).\]

    We now show concentration of each quantity
    $\kappa_s$ around its mean. The argument is
    similar to \cite[Theorem 3.9]{rudelsonvershynin}, and we omit some details.
    Write
    \[\kappa_s=\left\|\sum_{r=1}^n \A_r\A_r^*-\I\right\|_{K_s}\]
    where $\|\M\|_{K_s}=\sup_{\x \in K_s} |\x^*\M\x|$.
    Let $\A'$ be an independent copy of $\A$ and define
    \[\gamma_s=\left\|\sum_{r=1}^n \A_r\A_r^*-(\A_r')(\A_r')^*\right\|_{K_s}.\]
    Then by the same arguments as \cite[Theorem 3.9]{rudelsonvershynin},
    for any $t>0$,
    \begin{equation}
        \P[\kappa_s \geq 2\E[\kappa_s]+t] \leq 2\P[\gamma_s \geq t], \quad 
        \E[\gamma_s] \leq 2\E[\kappa_s] \leq 2E(p,n,s).\label{eq:sym1}
    \end{equation}
    From (\ref{eq:Anormbound}), we have
    \[\|\A_r\A_r^*\|_{K_s} \leq B^2\]
    for every $r$. Then applying \cite[Theorem 3.8]{rudelsonvershynin},
    for any integers $l \geq q$, any $r>0$, and some constants $C_1,C_2>0$,
    \[\P[\gamma_s \geq 8q\E[\gamma_s]+2B^2l+r]
    \leq (C_1/q)^l+2\exp\left(-\frac{r^2}{C_2q\E[\gamma_s]^2}\right).\]
    Let us assume without loss of generality $C_1 \geq 1/e$ and set
    $l=[2eC_1(\log n)(\log p)^3]$, $q=[eC_1]$, and
    $r=2\sqrt{C_2q} \cdot 2E(p,n,s) \cdot t\sqrt{\log p}$, where $[\cdot]$
    denotes the integer part. Then combining this with (\ref{eq:sym1}),
    we get for some constants $C,t_0>0$ and all $t>t_0$ that
    \[\P[\kappa_s \geq C\sqrt{t} \cdot E(p,n,s) \sqrt{\log p}]
    \leq e^{-2(\log n)(\log p)^3}+2e^{-4t(\log p)}.\]
    Setting $\rho(s)=Cst(\log p)^8(\log n)/n$ for a sufficiently large
    constant $C>0$, this yields
    \[\P[\kappa_s \geq 2\sqrt{\rho(s)}+\rho(s)] \leq e^{-2(\log n)(\log p)^3}
    +2e^{-4t(\log p)}.\]
    The result follows from taking a union bound over $s=1,\ldots,|E|$, and
    noting $|E| \leq 2p$ and 
    \[2p\left(e^{-2(\log n)(\log p)^3}+2e^{-4t(\log p)}\right)
    \leq e^{-(\log n)(\log p)^3}+p^{-t}\]
    for all $t>t_0$ and sufficiently large $t_0>0$.
\end{proof}

\section{RMSE for optimal parameter tuning}\label{appendix:bestachieved}
We report here the best-achieved RMSE, rather than RMSE for cross-validated
selection of tuning parameters, corresponding to Tables \ref{tab:spike},
\ref{tab:wave}, \ref{tab:shepplogan}, \ref{tab:brain}, and \ref{tab:XCAT}. For
the 1-D signals, we performed 20 independent simulations, and we report also the
standard deviations across these 20 simulations.

\begin{table}[H]
\begin{tabular}{l|c||ccccccccc}
$n/p$ & & $\sigma=0$ & $\sigma=1$ & $\sigma=2$ & $\sigma=3$ & $\sigma=4$ & $\sigma=5$ & $\sigma=6$ & $\sigma=7$ \\
\hline
\multirow{4}{2em}{10\%} & ITALE & {\bf 0.000} & {\bf 0.011} & {\bf 0.050} & {\bf 0.081} & {\bf 0.115} & {\bf 0.138} & {\bf 0.177} & {\bf 0.198} \\
&  & (0.000) & (0.004) & (0.024) & (0.023) & (0.033) & (0.034) & (0.031) & (0.025) \\
& TV & {\bf 0.000} & 0.045 & 0.086 & 0.122 & 0.152 & 0.176 & 0.193 & 0.206 \\
&  & (0.000) & (0.011) & (0.017) & (0.018) & (0.017) & (0.016) & (0.017) & (0.015) \\
\hline
\multirow{4}{2em}{15\%} & ITALE & {\bf 0.000} & {\bf 0.008} & {\bf 0.019} & {\bf 0.042} & {\bf 0.069} & {\bf 0.091} & {\bf 0.114} & {\bf 0.133} \\
&  & (0.000) & (0.002) & (0.007) & (0.016) & (0.023) & (0.029) & (0.028) & (0.029) \\
& TV & {\bf 0.000} & 0.029 & 0.057 & 0.085 & 0.109 & 0.130 & 0.149 & 0.165 \\
&  & (0.000) & (0.006) & (0.011) & (0.016) & (0.020) & (0.020) & (0.020) & (0.020) \\
\hline
\multirow{4}{2em}{20\%} & ITALE & {\bf 0.000} & {\bf 0.007} & {\bf 0.013} & {\bf 0.028} & {\bf 0.049} & {\bf 0.070} & {\bf 0.090} & {\bf 0.102} \\
&  & (0.000) & (0.002) & (0.003) & (0.012) & (0.019) & (0.019) & (0.017) & (0.020) \\
& TV & {\bf 0.000} & 0.022 & 0.044 & 0.066 & 0.087 & 0.107 & 0.126 & 0.143 \\
&  & (0.000) & (0.002) & (0.004) & (0.006) & (0.008) & (0.009) & (0.010) & (0.010) \\
\hline
\multirow{4}{2em}{30\%} & ITALE & {\bf 0.000} & {\bf 0.006} & {\bf 0.012} & {\bf 0.019} & {\bf 0.029} & {\bf 0.045} & {\bf 0.061} & {\bf 0.075} \\
&  & (0.000) & (0.001) & (0.003) & (0.006) & (0.009) & (0.014) & (0.011) & (0.013) \\
& TV & {\bf 0.000} & 0.017 & 0.034 & 0.051 & 0.068 & 0.085 & 0.101 & 0.116 \\
&  & (0.000) & (0.001) & (0.003) & (0.004) & (0.006) & (0.007) & (0.008) & (0.009) \\
\hline
\multirow{4}{2em}{40\%} & ITALE & {\bf 0.000} & {\bf 0.005} & {\bf 0.010} & {\bf 0.015} & {\bf 0.024} & {\bf 0.038} & {\bf 0.048} & {\bf 0.062} \\
&  & (0.000) & (0.001) & (0.002) & (0.003) & (0.007) & (0.012) & (0.015) & (0.016) \\
& TV & {\bf 0.000} & 0.014 & 0.028 & 0.042 & 0.056 & 0.070 & 0.084 & 0.097 \\
&  & (0.000) & (0.001) & (0.003) & (0.004) & (0.006) & (0.007) & (0.008) & (0.010) \\
\hline
\multirow{4}{2em}{50\%} & ITALE & {\bf 0.000} & {\bf 0.005} & {\bf 0.009} & {\bf 0.014} & {\bf 0.021} & {\bf 0.029} & {\bf 0.036} & {\bf 0.047} \\
&  & (0.000) & (0.001) & (0.002) & (0.003) & (0.006) & (0.011) & (0.013) & (0.013) \\
& TV & {\bf 0.000} & 0.013 & 0.025 & 0.038 & 0.050 & 0.063 & 0.074 & 0.086 \\
&  & (0.000) & (0.002) & (0.003) & (0.005) & (0.006) & (0.008) & (0.008) & (0.009) \\
\hline
\end{tabular}

\caption{Mean and standard deviation of best-achieved RMSE for the 1-D spike
signal across 20 simulations}
\end{table}

\newpage

\begin{table}[H]
\begin{tabular}{l|c||ccccccccc}
$n/p$ & & $\sigma=0$ & $\sigma=1$ & $\sigma=2$ & $\sigma=3$ & $\sigma=4$ & $\sigma=5$ & $\sigma=6$ & $\sigma=7$ \\
\hline
\multirow{4}{2em}{10\%} & ITALE & 0.019 & {\bf 0.030} & 0.075 & 0.126 & 0.155 & 0.203 & 0.233 & 0.256 \\
&  & (0.081) & (0.066) & (0.055) & (0.057) & (0.064) & (0.059) & (0.070) & (0.068) \\
& TV & {\bf 0.000} & 0.031 & {\bf 0.062} & {\bf 0.090} & {\bf 0.114} & {\bf 0.137} & {\bf 0.157} & {\bf 0.176} \\
&  & (0.000) & (0.005) & (0.011) & (0.016) & (0.018) & (0.021) & (0.022) & (0.024) \\
\hline
\multirow{4}{2em}{15\%} & ITALE & {\bf 0.000} & {\bf 0.008} & {\bf 0.022} & {\bf 0.052} & {\bf 0.083} & {\bf 0.105} & 0.134 & 0.154 \\
&  & (0.000) & (0.002) & (0.011) & (0.023) & (0.027) & (0.030) & (0.027) & (0.035) \\
& TV & {\bf 0.000} & 0.022 & 0.044 & 0.066 & 0.087 & 0.106 & {\bf 0.124} & {\bf 0.141} \\
&  & (0.000) & (0.003) & (0.006) & (0.009) & (0.011) & (0.013) & (0.014) & (0.016) \\
\hline
\multirow{4}{2em}{20\%} & ITALE & {\bf 0.000} & {\bf 0.006} & {\bf 0.015} & {\bf 0.032} & {\bf 0.050} & {\bf 0.072} & {\bf 0.093} & {\bf 0.109} \\
&  & (0.000) & (0.001) & (0.006) & (0.011) & (0.016) & (0.023) & (0.027) & (0.027) \\
& TV & {\bf 0.000} & 0.018 & 0.036 & 0.054 & 0.071 & 0.088 & 0.104 & 0.118 \\
&  & (0.000) & (0.002) & (0.004) & (0.007) & (0.009) & (0.011) & (0.012) & (0.013) \\
\hline
\multirow{4}{2em}{30\%} & ITALE & {\bf 0.000} & {\bf 0.006} & {\bf 0.011} & {\bf 0.017} & {\bf 0.032} & {\bf 0.046} & {\bf 0.062} & {\bf 0.077} \\
&  & (0.000) & (0.001) & (0.002) & (0.004) & (0.010) & (0.012) & (0.015) & (0.015) \\
& TV & {\bf 0.000} & 0.014 & 0.027 & 0.041 & 0.055 & 0.068 & 0.082 & 0.095 \\
&  & (0.000) & (0.001) & (0.003) & (0.004) & (0.005) & (0.007) & (0.008) & (0.010) \\
\hline
\multirow{4}{2em}{40\%} & ITALE & {\bf 0.000} & {\bf 0.005} & {\bf 0.010} & {\bf 0.016} & {\bf 0.024} & {\bf 0.033} & {\bf 0.046} & {\bf 0.057} \\
&  & (0.000) & (0.001) & (0.002) & (0.004) & (0.009) & (0.012) & (0.014) & (0.017) \\
& TV & {\bf 0.000} & 0.012 & 0.024 & 0.036 & 0.048 & 0.060 & 0.072 & 0.083 \\
&  & (0.000) & (0.001) & (0.002) & (0.003) & (0.005) & (0.006) & (0.007) & (0.008) \\
\hline
\multirow{4}{2em}{50\%} & ITALE & {\bf 0.000} & {\bf 0.004} & {\bf 0.009} & {\bf 0.015} & {\bf 0.022} & {\bf 0.030} & {\bf 0.038} & {\bf 0.049} \\
&  & (0.000) & (0.001) & (0.002) & (0.005) & (0.009) & (0.011) & (0.012) & (0.018) \\
& TV & {\bf 0.000} & 0.011 & 0.021 & 0.032 & 0.042 & 0.053 & 0.063 & 0.073 \\
&  & (0.000) & (0.001) & (0.003) & (0.004) & (0.005) & (0.007) & (0.008) & (0.009) \\
\hline
\end{tabular}

\caption{Mean and standard deviation of best-achieved RMSE for the 1-D wave
signal across 20 simulations}
\end{table}

\begin{table}[H]
\begin{tabular}{l|c||cccccccc}
$n/p$ & & $\sigma=0$ & $\sigma=4$ & $\sigma=8$ & $\sigma=12$ & $\sigma=16$ & $\sigma=20$ & $\sigma=24$ & $\sigma=28$ \\
\hline
\multirow{2}{2em}{10\%} & ITALE & {\bf 0.000} & {\bf 0.004} & {\bf 0.012} & {\bf 0.016} & {\bf 0.023} & {\bf 0.037} & {\bf 0.046} & 0.067 \\
& TV & 0.002 & 0.011 & 0.021 & 0.030 & 0.038 & 0.049 & 0.056 & {\bf 0.061} \\
\hline
\multirow{2}{2em}{15\%} & ITALE & {\bf 0.000} & {\bf 0.003} & {\bf 0.010} & {\bf 0.013} & {\bf 0.017} & {\bf 0.025} & {\bf 0.035} & {\bf 0.044} \\
& TV & 0.001 & 0.008 & 0.016 & 0.025 & 0.030 & 0.038 & 0.047 & 0.052 \\
\hline
\multirow{2}{2em}{20\%} & ITALE & {\bf 0.000} & {\bf 0.003} & {\bf 0.008} & {\bf 0.011} & {\bf 0.016} & {\bf 0.018} & {\bf 0.026} & {\bf 0.035} \\
& TV & 0.001 & 0.007 & 0.014 & 0.021 & 0.027 & 0.032 & 0.039 & 0.045 \\
\hline
\multirow{2}{2em}{30\%} & ITALE & {\bf 0.000} & {\bf 0.002} & {\bf 0.005} & {\bf 0.011} & {\bf 0.013} & {\bf 0.015} & {\bf 0.018} & {\bf 0.023} \\
& TV & 0.002 & 0.006 & 0.011 & 0.017 & 0.022 & 0.027 & 0.032 & 0.036 \\
\hline
\multirow{2}{2em}{40\%} & ITALE & {\bf 0.000} & {\bf 0.001} & {\bf 0.005} & {\bf 0.009} & {\bf 0.011} & {\bf 0.013} & {\bf 0.016} & {\bf 0.018} \\
& TV & 0.001 & 0.005 & 0.010 & 0.014 & 0.019 & 0.023 & 0.026 & 0.031 \\
\hline
\multirow{2}{2em}{50\%} & ITALE & {\bf 0.000} & {\bf 0.001} & {\bf 0.004} & {\bf 0.008} & {\bf 0.010} & {\bf 0.012} & {\bf 0.013} & {\bf 0.015} \\
& TV & 0.001 & 0.005 & 0.009 & 0.013 & 0.017 & 0.021 & 0.025 & 0.029 \\
\hline
\end{tabular}

\caption{Best-achieved RMSE for the Shepp-Logan phantom}
\end{table}

\begin{table}[H]
\begin{tabular}{l|c||cccccccc}
$n/p$ & & $\sigma=0$ & $\sigma=8$ & $\sigma=16$ & $\sigma=24$ & $\sigma=32$ & $\sigma=40$ & $\sigma=48$ & $\sigma=56$ \\
\hline
\multirow{2}{2em}{10\%} & ITALE & {\bf 0.000} & {\bf 0.002} & {\bf 0.010} & {\bf 0.028} & {\bf 0.042} & {\bf 0.058} & {\bf 0.077} & 0.093 \\
& TV & 0.002 & 0.014 & 0.028 & 0.041 & 0.054 & 0.064 & {\bf 0.077} & {\bf 0.085} \\
\hline
\multirow{2}{2em}{15\%} & ITALE & {\bf 0.000} & {\bf 0.001} & {\bf 0.007} & {\bf 0.017} & {\bf 0.030} & {\bf 0.043} & {\bf 0.057} & 0.075 \\
& TV & 0.001 & 0.011 & 0.022 & 0.032 & 0.043 & 0.053 & 0.063 & {\bf 0.071} \\
\hline
\multirow{2}{2em}{20\%} & ITALE & {\bf 0.000} & {\bf 0.001} & {\bf 0.005} & {\bf 0.012} & {\bf 0.025} & {\bf 0.034} & {\bf 0.047} & {\bf 0.054} \\
& TV & 0.001 & 0.009 & 0.019 & 0.028 & 0.037 & 0.044 & 0.055 & 0.061 \\
\hline
\multirow{2}{2em}{30\%} & ITALE & {\bf 0.000} & {\bf 0.001} & {\bf 0.002} & {\bf 0.007} & {\bf 0.014} & {\bf 0.024} & {\bf 0.033} & {\bf 0.043} \\
& TV & 0.001 & 0.007 & 0.016 & 0.022 & 0.030 & 0.038 & 0.044 & 0.052 \\
\hline
\multirow{2}{2em}{40\%} & ITALE & {\bf 0.000} & {\bf 0.000} & {\bf 0.002} & {\bf 0.005} & {\bf 0.010} & {\bf 0.018} & {\bf 0.026} & {\bf 0.032} \\
& TV & 0.001 & 0.007 & 0.013 & 0.019 & 0.026 & 0.032 & 0.039 & 0.043 \\
\hline
\multirow{2}{2em}{50\%} & ITALE & {\bf 0.000} & {\bf 0.001} & {\bf 0.002} & {\bf 0.004} & {\bf 0.007} & {\bf 0.012} & {\bf 0.019} & {\bf 0.027} \\
& TV & 0.001 & 0.006 & 0.011 & 0.017 & 0.023 & 0.029 & 0.034 & 0.040 \\
\hline
\end{tabular}

\caption{Best-achieved RMSE for the brain phantom}
\end{table}

\begin{table}[H]
\begin{tabular}{l|c||cccccccc}
$n/p$ & & $\sigma=0$ & $\sigma=4$ & $\sigma=8$ & $\sigma=12$ & $\sigma=16$ & $\sigma=20$ & $\sigma=24$ & $\sigma=28$ \\
\hline
\multirow{2}{2em}{10\%} & ITALE & {\bf 0.002} & 0.032 & 0.043 & 0.068 & 0.075 & 0.084 & 0.093 & 0.101 \\
& TV & 0.006 & {\bf 0.019} & {\bf 0.032} & {\bf 0.044} & {\bf 0.053} & {\bf 0.061} & {\bf 0.068} & {\bf 0.072} \\
\hline
\multirow{2}{2em}{15\%} & ITALE & {\bf 0.002} & {\bf 0.007} & {\bf 0.017} & {\bf 0.029} & 0.044 & 0.067 & 0.076 & 0.084 \\
& TV & 0.003 & 0.014 & 0.024 & 0.034 & {\bf 0.043} & {\bf 0.050} & {\bf 0.056} & {\bf 0.061} \\
\hline
\multirow{2}{2em}{20\%} & ITALE & {\bf 0.002} & {\bf 0.005} & {\bf 0.014} & {\bf 0.022} & {\bf 0.032} & {\bf 0.041} & 0.052 & 0.069 \\
& TV & {\bf 0.002} & 0.011 & 0.020 & 0.028 & 0.037 & 0.043 & {\bf 0.049} & {\bf 0.056} \\
\hline
\multirow{2}{2em}{30\%} & ITALE & {\bf 0.002} & {\bf 0.004} & {\bf 0.011} & {\bf 0.017} & {\bf 0.024} & {\bf 0.030} & {\bf 0.038} & 0.049 \\
& TV & {\bf 0.002} & 0.008 & 0.016 & 0.023 & 0.029 & 0.035 & 0.041 & {\bf 0.046} \\
\hline
\multirow{2}{2em}{40\%} & ITALE & 0.002 & {\bf 0.003} & {\bf 0.008} & {\bf 0.014} & {\bf 0.020} & {\bf 0.026} & {\bf 0.032} & {\bf 0.038} \\
& TV & {\bf 0.001} & 0.007 & 0.014 & 0.020 & 0.025 & 0.031 & 0.036 & 0.041 \\
\hline
\multirow{2}{2em}{50\%} & ITALE & {\bf 0.001} & {\bf 0.003} & {\bf 0.007} & {\bf 0.013} & {\bf 0.017} & {\bf 0.023} & {\bf 0.028} & {\bf 0.033} \\
& TV & {\bf 0.001} & 0.006 & 0.012 & 0.018 & 0.023 & 0.028 & 0.033 & 0.038 \\
\hline
\end{tabular}

\caption{Best-achieved RMSE for the XCAT chest slice phantom}
\end{table}

\bibliographystyle{alpha}
\bibliography{references.bib}

\end{document}